\documentclass[runningheads]{llncs}
\usepackage[T1]{fontenc}
\usepackage{graphicx,verbatim}
\usepackage{enumitem}
\usepackage{amsmath}
\usepackage{amsfonts}
\usepackage{stmaryrd}
\usepackage{makecell} % add this in the preamble  
\usepackage{booktabs}
\usepackage[table]{xcolor}
\usepackage{graphicx}
\usepackage{adjustbox}
\usepackage{url}
\usepackage[colorlinks=true]{hyperref}

\title{IMILIA: interpretable multiple instance learning for inflammation prediction in IBD from H\&E whole slide images}
\titlerunning{IMILIA}

 \author{Thalyssa Baiocco-Rodrigues$^{1, *}$, Antoine Olivier$^{1, *}$, Reda Belbahri$^1$, Thomas Duboudin$^1$, Pierre-Antoine Bannier$^1$, Benjamin Adjadj$^1$, Katharina Von Loga$^1$, Nathan Noiry$^1$, Maxime Touzot$^1$, Hector Roux de Bezieux$^1$} 
\authorrunning{\authorrunning{T. Baiocco-Rodrigues et al.}}
\institute{
    {$^1$ Owkin, Inc} \\
    {$^*$ Equal contribution.} \\
    {Correspondance to \texttt{\{thalyssa.baiocco-rodrigues, antoine.olivier\}@owkin.com}}
}

\begin{document}

\maketitle

\begin{abstract}
As the therapeutic target for Inflammatory Bowel Disease (IBD) shifts toward histologic remission, the accurate assessment of microscopic inflammation has become increasingly central for evaluating disease activity and response to treatment.
In this work, we introduce IMILIA (Interpretable Multiple Instance Learning for Inflammation Analysis), an end-to-end framework designed for the prediction of inflammation presence in IBD digitized slides stained with hematoxylin and eosin (H\&E), followed by the automated computation of markers characterizing tissue regions driving the predictions. IMILIA is composed of an inflammation prediction module, consisting of a Multiple Instance Learning (MIL) model, and an interpretability module, divided in two blocks: HistoPLUS, for cell instance detection, segmentation and classification; and EpiSeg, for epithelium segmentation. 
IMILIA achieves a cross-validation ROC-AUC of 0.83 on the discovery cohort, and a ROC-AUC of 0.99 and 0.84 on two external validation cohorts.
The interpretability module yields biologically consistent insights: tiles with higher predicted scores show increased densities of immune cells (lymphocytes, plasmocytes, neutrophils and eosinophils), whereas lower-scored tiles predominantly contain normal epithelial cells. Notably, these patterns were consistent across all datasets. 
Code and models to partially replicate the results on the public IBDColEpi dataset can be found at \url{https://github.com/owkin/imilia}.
\end{abstract}


\section{Introduction}

\paragraph{Background.} Inflammatory Bowel Disease (IBD) is a lifelong, chronic inflammatory disorder of the gastrointestinal tract, manifesting as Crohn’s disease (CD) or ulcerative colitis (UC) \cite{Graham2020}. IBD places significant strain on healthcare systems \cite{burisch2025}, affecting nearly 7 million patients worldwide \cite{Feuerstein2014,Molodecky2012,Ng2017}, with established high burdens in Europe (2 million) and North America (1.5 million). While clinical symptoms, ranging from abdominal pain to extra-intestinal manifestations, are primary indicators, they are often non-specific as the disease course is characterized by periods of active flare-ups interspersed with remission. 
Consequently, diagnosis and management heavily rely on a combination of endoscopic and histological assessments \cite{angyal2025}. Histopathological assessment of Hematoxylin and Eosin (H\&E) stained slides is critical not only to confirm the diagnosis and exclude differentials (\emph{e.g.}, granulomatosis) but also to grade severity. Interestingly, patients without lesions at endoscopy (endoscopic remission) might still harbor persistent microscopic inflammation associated with disease progression \cite{IACUCCI2021,Marchal-Bressenot2017}. This has led to a shift in the therapeutic goal in IBD, beyond symptom control and towards ``histological healing'' \cite{angyal2025,Villanacci2025}, as persistent microscopic inflammation is a strong predictor of relapse and colorectal cancer risk. 
However, the standard manual review of biopsy slides is time-consuming and dependent on the expertise of the reviewing pathologist \cite{Feakins2023,Geboes2000}. As the volume of biopsies increases, the need for automated decision-support tools becomes critical to ensure consistency and efficiency in patient care. 

\paragraph{Related work.} Modern computational pathology frameworks often rely on the combination of a feature extraction step using a foundation model (FM), and a downstream multiple instance learning (MIL) model. The FM maps small areas of tissue (or tiles) to a lower-dimensional representation space, and the MIL model further combines those representations to derive a slide-level prediction. While recent advances on FM development have yielded spectacular progress across a variety of downstream tasks \cite{Chen2024}, the majority of applications have remained limited to oncology.
In the specific domain of IBD, recent research has focused on automating histological grading, with models developed to predict established severity scores such as the NHI, Geboes Score, or Robarts Histopathology Index \cite{IACUCCI2023,Peyrin-Biroulet2024,Rubin2024,Plattner2025,Rymarczyk2023}. Beyond simple classification, efforts have also explored correlating histological features with endoscopic findings \cite{Divicenzo2024,Elta1987} and quantifying specific cellular populations at sample level \cite{Reigle2024}. In \cite{Ding2025} a graph neural network is introduced to predict WSI-level ulcer presence from H\&E WSIs. Closely related to our work, \cite{mokhtari24a} propose a MIL-based framework to predict disease type, macroscopic tissue appearance and endoscopic scores, while combining it with a cell detection model to derive interpretable insights. While these studies share similarities with our work, they differ mostly on the extent of the external validation, sometimes reporting results in cross-validation only, and the interpretability, often limited to generating heatmaps, remains mostly qualitative and lacks quantitative analysis.

\paragraph{Contributions.} In this work, we introduce IMILIA (Interpretable Multiple Instance Learning for Inflammation Analysis), an end-to-end interpretable MIL framework designed for the binary classification of IBD histology slides (inflamed vs. non-inflamed). We summarize our contributions as follows:
\begin{itemize}[label=$\bullet$] 
    \item We provide an extensive validation of IMILIA, using one large discovery cohort (over 3000 patients) and two external validation cohorts. To the best of our knowledge, this is the only validation of a deep learning model for IBD inflammation prediction on external datasets.
    \item We also provide a quantitative validation of the interpretability module, composed of HistoPLUS and EpiSeg in the context of IBD. Their combination allows to automatically derive advanced and potentially novel markers of the presence of inflammation, such as the localization of immune cells within the epithelium.
\end{itemize}
We release the code to reproduce the results on the public dataset with the Chowder and EpiSeg models (including trained Chowder) at \url{https://github.com/owkin/imilia}.

\section{IMILIA}

\begin{figure}[ht!]
 % Caption and label go in the first argument and the figure contents
 % go in the second argument
  \includegraphics[width=0.92\linewidth]{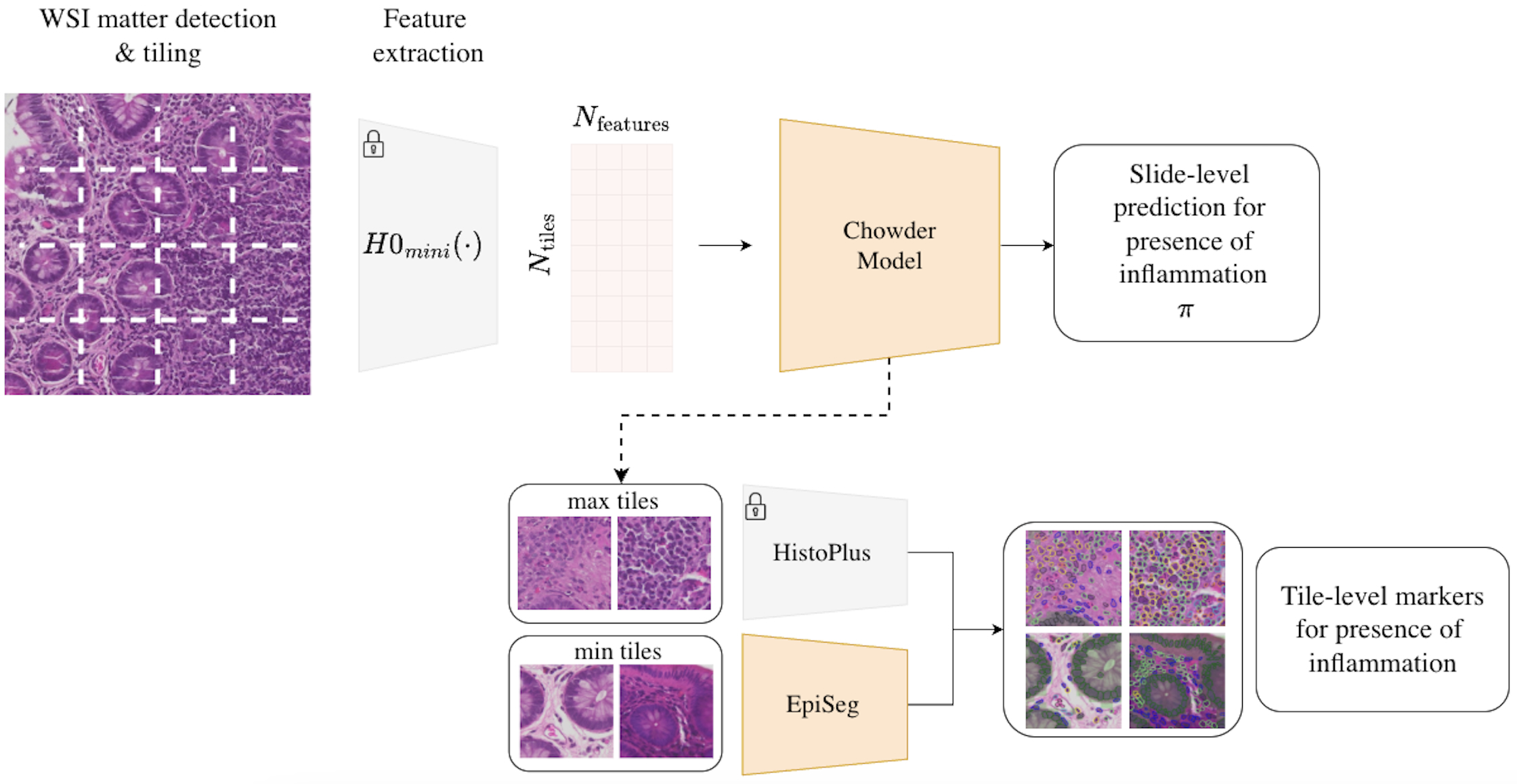}
  \caption{IMILIA: Interpretable Multiple Instance Learning for Inflammation Analysis. A multiple instance learning model is combined with two interpretability blocks to predict the presence of inflammation from H\&E slides in IBD, and derive quantitative biological markers driving the predictions. Beige colors denotes models specifically trained for this paper.}
  \label{fig:imilia_graphical_abstract}
\end{figure}

In this section, we introduce IMILIA and explain how a MIL model can be combined with an interpretability module to predict inflammation from H\&E WSI. An overview of the method is represented in Figure \ref{fig:imilia_graphical_abstract}.

\paragraph{WSI preprocessing.} An in-house segmentation network is used to detect regions with relevant tissue, excluding background and acquisition artifacts. The tissue regions are then tessellated into non-overlapping tiles of fixed dimensions. %This results in a bag of tiles for each slide, whose size varies based on the tissue surface area. More details about the pre-processing can be found in the appendix.

\paragraph{Feature extraction.} Throughout this study, and for consistency, we use H0-mini \cite{filiot2025} as a feature extractor to transform tissue tiles into low-dimensional representations. We use H0-mini because it offers one of the best trade-offs between performance and computational efficiency, while exhibiting strong robustness to variations in scanning and staining conditions.

\paragraph{Chowder.} Similar to other MIL architectures \cite{ilse2018}, Chowder \cite{courtiol2020} identifies a WSI to a collection of tiles. Following the feature extraction step, each slide is therefore represented by a matrix $W \in \mathbb{R}^{n_{tiles} \times d}$, where $n_{tiles}$ is the number of tiles in the slide and $d$ is the dimension of the embeddings. A first 1D convolution layer maps $W$ to a score vector $S \in \mathbb{R}^{n_{tiles} \times 1}$. Then, extreme tiles are selected by keeping only the $r$ max and min scores in S, resulting in a vector $S_{\text{extr}} \in \mathbb{R}^{(2 \times r) \times 1}$. A final MLP is then applied to produce the model’s prediction. In this study, we use a multi-channel version of Chowder, where the first 1D convolution layer is replaced by K 1D convolution layers that operate in parallel, resulting in a vector $S_{\text{extr}} \in \mathbb{R}^{(2 \times r \times K) \times 1}$.

\paragraph{Selection of extreme tiles.} As a by-product of its training process, Chowder learns tile score representations $S$, where tiles with maximal scores are associated with a positive prediction and tiles with minimal scores are associated with a negative prediction. In the rest of this paper, such tiles are referred to as max and min tiles. At inference, scores are computed for all slides, allowing to compare the content of tiles providing positive evidence of inflammation (higher scores) versus tiles providing negative evidence (lower scores).

\paragraph{HistoPlus \cite{adjadj2025}.} As a first block of the interpretability module, we use HistoPLUS , a cell detection model trained on a pancancer dataset covering 6 indications, and supporting 13 different cell types, and consisting of a CellViT \cite{Horst2024} model. While we note that HistoPLUS was originally trained on oncology WSIs, we provide quantitative evidence in Section \ref{sec:results} of its transfer performance to IBD. Following the extreme tiles selection, HistoPLUS can be applied to derive tile-level cell types predictions. 

\paragraph{EpiSeg.} EpiSeg is the second block of the interpretability module, an epithelium segmentation model, trained on top of H0-mini's representations of patch tokens. Given a tile of size $W \times H$, and a patch size $P$, the tile is divided into $W / P \times H / P$ small patches. Following the H0-mini inference, each patch is then represented by a $d$-dimensional vector, or patch embedding. A logistic regression is then trained to predict the presence of epithelium at patch level, resulting in segmentation maps of size $W / P \times H / P$.
Similar to HistoPLUS, EpiSeg can then be applied on the extreme tiles of Chowder to produce coarse segmentation maps of the presence of epithelium.

\section{Data}

For this study, we are using three datasets: the SPARC IBD \cite{Sparcibd2021} dataset as our discovery cohort, the FINBB and IBDColEpi \cite{Pettersen2022} datasets as our two external validation datasets for the inflammation prediction module. IBDColEpi was used as training for the EpiSeg model.

\paragraph{SPARC IBD.} The Study of a Prospective Adult Research Cohort with IBD (SPARC IBD) is a multicentered longitudinal study of adult IBD patients. We included 3322 H\&E slides obtained from intestinal mucosal biopsies of patients to compose our discovery cohort, among which $50.2\%$ correspond to patients diagnosed with CD and $24.8\%$ with UC (missing diagnostic information for other samples). $67.8\%$ of samples were extracted from the colon and $26.9\%$ from the ileum (missing location information for other samples). Ground-truth labels for inflammation presence were derived from the macroscopic tissue appearance of slides, assessed by a pathologist, and originally categorized into 1) normal, 2) possible inflammation and 3) severe inflammation, with presence of erosion or ulcers. We grouped categories 2 and 3 to compose the “inflamed” class (the positive class used for training). The prevalence of inflamed samples is $31\%$ ($N=1022$).

\paragraph{FINBB.} This dataset has been provided by the Finnish Biobank Cooperative (FINBB) and includes comprehensive multimodal data collected from Helsinki Biobank and Finnish Clinical Biobank Tampere. We used a total of 314 H\&E slides from this dataset as external validation, among which $59.2\%$ corresponded to patients diagnosed with UC and $40.8\%$ with CD. $76.4\%$ samples were extracted from colon, $20\%$ from ileum and $1.6\%$ from the intersection ileum/colon (other samples were classified as “unclear”). As for SPARC IBD, labels were derived from the macroscopic tissue appearance of slides, categorized into 1) normal and 2) inflamed. We note the prevalence shift compared to the discovery cohort, with inflamed samples accounting for $95\%$ of the population.

\paragraph{IBDColEpi.} The dataset consists of 140 H\&E WSIs from biopsies of colonic mucosa of active and inactive IBD with pixel-level annotation of the epithelium. ``Active'' is defined as the presence of intraepithelial granulocytes in one or more locations in the biopsies. Still, the changes may be focal, hence the majority of the epithelium may still lack intraepithelial granulocytes or other signs of active disease (crypt abscesses, granulation tissue, etc.). For external validation, we use 132 H\&E slides (8 were excluded due to insufficient tissue area), with a prevalence of $41\%$ for positive (active disease) samples. No information is provided regarding categorization of samples into UC vs CD. IBDColEpi is publicly available (\url{https://www.kaggle.com/datasets/henrikpe/251-he-cd3-wsis-annotated-epithelium-ibdcolepi}).

\section{Experimental setup.} 
\label{sec:exp-setup}

\paragraph{Chowder implementation.} For the MIL model, a multi-channel Chowder model with 5 channels was implemented. The number of selected extreme tiles per channel is 50, 25 with maximal scores and 25 with minimal scores, resulting in 250 extreme tiles. A final MLP with 2 hidden layers maps the tiles scores to the final prediction. The model is trained on the SPARC IBD dataset with a 5-fold cross-validation, using the Adam optimizer, and dropout is applied to the linear layers with probability $0.5$. At inference, on the external validation cohorts, the 5 models are ensembled by averaging their predictions (including for the min and max tile scores). We give in Appendix \ref{app:hyperparameters} an extensive list of the hyperparameters used in the training process.

For each cohort, two subsets of $1000$ tiles, one with maximal and the other with minimal scores, are selected and processed by the two interpretability models, EpiSeg and HistoPLUS. %More details about the selection of these groups of tiles can be found in the appendix.

\paragraph{EpiSeg training.} We leverage the epithelium annotations in IBDColEpi, and start from images and their corresponding pixel-level segmentation masks of epithelium of the same size. We note that the $224 \times 224$ tiles used to train the MIL model are too small to correctly predict the presence of epithelium, therefore, and to increase the context, we trained EpiSeg on tiles of size $1022 \times 1022$, which simply corresponds to resampling the images available in the dataset to a spatial resolution of $0.5$ micrometer per pixel (mpp). H0-mini operates by dividing images into small patches of constant size $14 \times 14$. Following the forward pass in H0-mini, each patch is associated with an embedding $x_p$ of dimension $d=768$. To obtain lower resolution segmentation masks, we simply apply a 14 x 14 convolution filter with stride 14 and constant weights (and produce the equivalent of an average-pooling layer over patches). At this stage, each patch is associated with a continuous label $y_p$ which can be interpreted as the area of epithelium within this patch. A logistic regression is finally trained on pairs $(x_p, y_p)$. A 3-fold cross-validation is performed to optimize the $L2$-regularization parameter $C$ in the model, which yields an optimal value $C = 10^{-2}$ used to train the final model. We use the train / test split provided by the authors of the dataset to train and evaluate our model. Illustrating samples can be found in Figure \ref{fig:episeg_gt_pred}.

\paragraph{EpiSeg inference on extreme tiles.} At inference time, for consistency in the patches representations by H0-mini, we ensure images have the same size $1022 \times 1022$ as was used during the training of EpiSeg. We thus extract expanded versions of the minimal and maximal tiles to match this size, with the tile at its center. The expanded image is used for EpiSeg inference and the final epithelium mask for the tile of interest is obtained by cropping the center of the EpiSeg’s output mask.

\paragraph{Interpretable features computation.} For each max and min tiles, HistoPLUS and EpiSeg were used to generate cell masks (one for each cell type) and epithelium masks, respectively. They are further combined to compute the density $\rho^c$ of several cell types $c$ within the epithelium region, as: 
\begin{equation}
\rho^c := \frac{\sum_{k=0}^{N^c}E(c_k)}{\sum_{j=0}^{H}\sum_{i=0}^{W}E(x_i,y_j) \times \text{mpp}_x \times \text{mpp}_y} 
\end{equation}
where, for a pixel location $(x, y)$, $E(x,y)$ is one if $(x, y)$ falls within the predicted epithelium region; $N^c$ is the total number of instances for the cell type $c$, $H$ and $W$ are the height and width of the epithelium mask, $c_k := (c_{k,x}, c_{k,y})$ is the centroid of the $k$-th cell instance, and $(\text{mpp}_x, \text{mpp}_y)$ is the pixel size. %This yields interpretable features in the format “N cells of type $c$ per um2”.

\section{Results}
\label{sec:results}

\paragraph{IMILIA shows strong cross-validation and transfer performance on the three cohorts.} IMILIA's classification module (Chowder model) shows a ROC-AUC of $0.83$ in cross-validation in the discovery dataset SPARC IBD. It shows a great transferability in external datasets, reaching an AUC of $0.84$ on FINBB. Remarkably, the model achieved near-perfect classification on the IBDColEpi dataset, with an AUC of $0.99$. We attribute this to the curated nature of the dataset, which likely represents distinct inflammatory phenotypes with minimal ``grey-zone'' cases.

\paragraph{HistoPLUS demonstrates robust transfer performance in a new therapeutic area, beyond oncology.}

To assess the generalizability of HistoPLUS beyond oncology, and its suitability for IBD, we validated it in 90 tiles extracted from the SPARC IBD dataset, annotated with $13291$ cells overall. Originally developed for oncology applications, the discrimination between cancer and healthy epithelial cells was part of HistoPLUS’ training. However, its training dataset may not adequately represent the full spectrum of epithelial morphologies, in particular in the heavily immune-infiltrated setting characteristic of IBD where epithelial cells frequently exhibit morphological irregularities that can resemble neoplastic features. Given these considerations, we elected to remap all predicted cancer cells to the epithelial class.

We use the same notation as in the HistoPLUS original paper, and denote by HistoVAL ($N=530$) the validation set of HistoPLUS in the oncology setting. Table \ref{tab:histoplus-classification-perf} and Table \ref{tab:histoplus-detection-perf} summarize the model’s performances, providing a direct comparison between SPARC IBD and HistoVAL. HistoPLUS exhibited consistent validation performances on IBD, achieving a detection quality score of $0.774$ ($95\% \text{CI}, [0.760; 0.789]$) on the IBD samples, and a segmentation quality score of $0.755$ ($95\% \text{CI}, [0.749; 0.761]$). For cell classification, HistoPLUS demonstrated good capabilities on epithelial cells, immune cells, endothelial cells and fibroblasts with no significant performance drop between oncology and IBD.

\begin{table}[htbp]
    \centering
    \caption{Cell classification performance of HistoPLUS in HistoVAL (oncology) and SPARC IBD. We report mean values of F1 scores and confidence intervals at 95\% level, obtained by bootstrapping with 1000 repeats.}%
    \label{tab:histoplus-classification-perf}
    \fontsize{8}{\baselineskip}\selectfont
     \begin{tabular}{lccccccc}
     \toprule
     \bfseries Dataset & \bfseries Epith. & \bfseries Lymph. & \bfseries Plasm. & \bfseries Eosin. & \bfseries Neutro. & \bfseries Endoth. & \bfseries Fibro.\\
     \midrule
     HistoVAL 
       & \makecell{0.42 \\ (0.28;0.54)}
       & \makecell{0.42 \\ (0.28;0.54)}
       & \makecell{0.48 \\ (0.43;0.53)}
       & \makecell{0.46 \\ (0.37;0.52)}
       & \makecell{0.24 \\ (0.18;0.31)}
       & \makecell{0.33 \\ (0.29;0.38)}
       & \makecell{0.38 \\ (0.35;0.40)} \\
     IBD
       & \makecell{0.70 \\ (0.67;0.73)}
       & \makecell{0.53 \\ (0.48;0.57)}
       & \makecell{0.52 \\ (0.46;0.57)}
       & \makecell{0.45 \\ (0.40;0.51)}
       & \makecell{0.36 \\ (0.29;0.42)}
       & \makecell{0.44 \\ (0.36;0.51)}
       & \makecell{0.33 \\ (0.27;0.39)} \\
     \bottomrule
     \end{tabular}
    
\end{table}

\paragraph{EpiSeg, a logistic regression trained on top of H0-mini’s representations can efficiently predict the presence of epithelium.} We use the test set of IBDColEPI to evaluate the logistic regression. It reaches an average precision score (defined as the area under the precision-recall curve) of $0.98$ to predict the presence of epithelium at patch level. Qualitative examples of EpiSeg’s predictions are displayed on Figure \ref{fig:episeg_gt_pred}, and the full precision-recall curve can be found in Appendix \ref{app:pr-curve}.

As an additional validation of EpiSeg, we found that HistoPLUS and EpiSeg show a strong agreement regarding the prediction of epithelium. The pearson correlation between the number of epithelial cells identified by HistoPLUS in a tile and the epithelium area from EpiSeg’s predictions in the same tile is 0.85 ($p < 10^{-8}$) for SPARC IBD, 0.74 ($p < 10^{-8}$) for FINBB and 0.83 ($p < 10^{-8}$) for IBDColEpi.

We also note that EpiSeg's performance provides strong evidence of the quality of H0-mini’s representations in the context of IBD, since a simple logistic regression model is able to learn how to discriminate patches corresponding to epithelium zones.

\begin{figure}[ht!]
 % Caption and label go in the first argument and the figure contents
 % go in the second argument
    \centering
    \begin{tabular}{ccc}
        \begin{minipage}{0.33\textwidth}
            \centering
            \includegraphics[width=\linewidth]{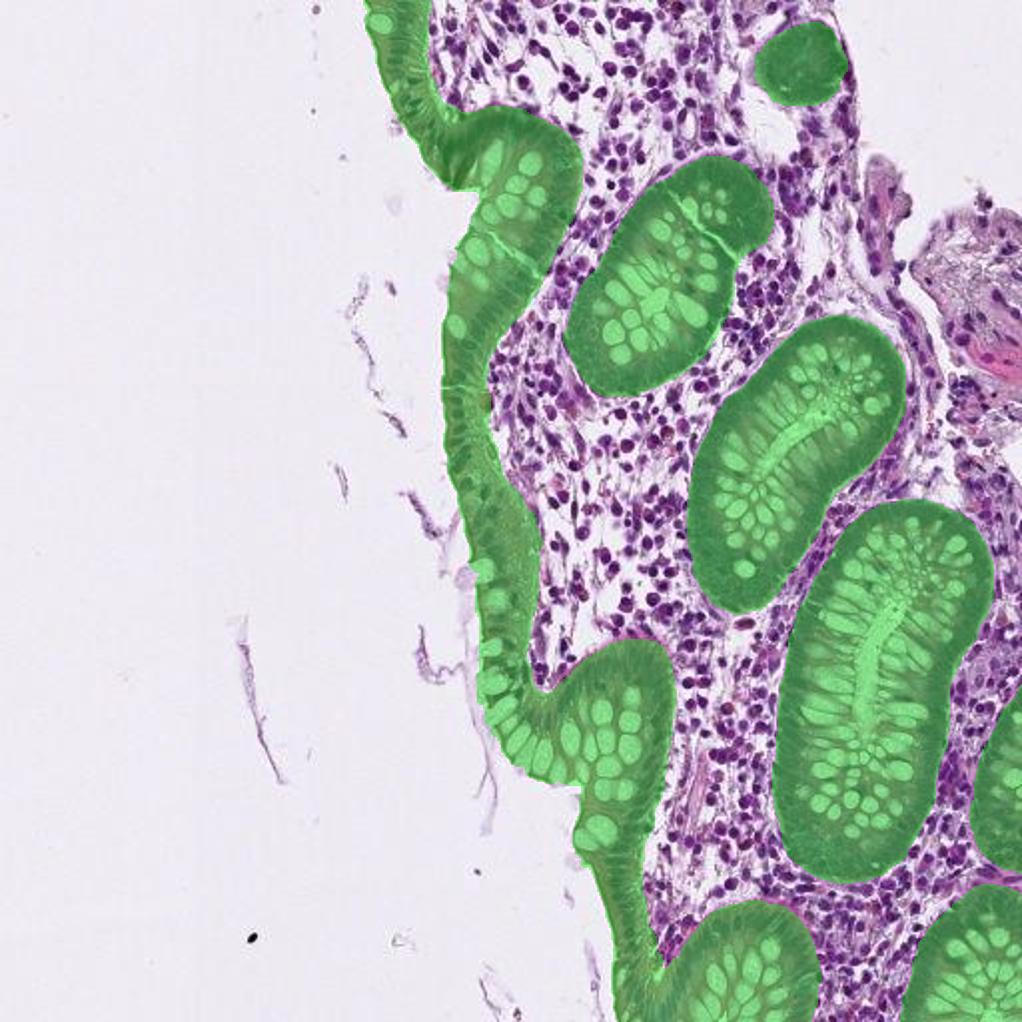}\\
            \includegraphics[width=\linewidth]{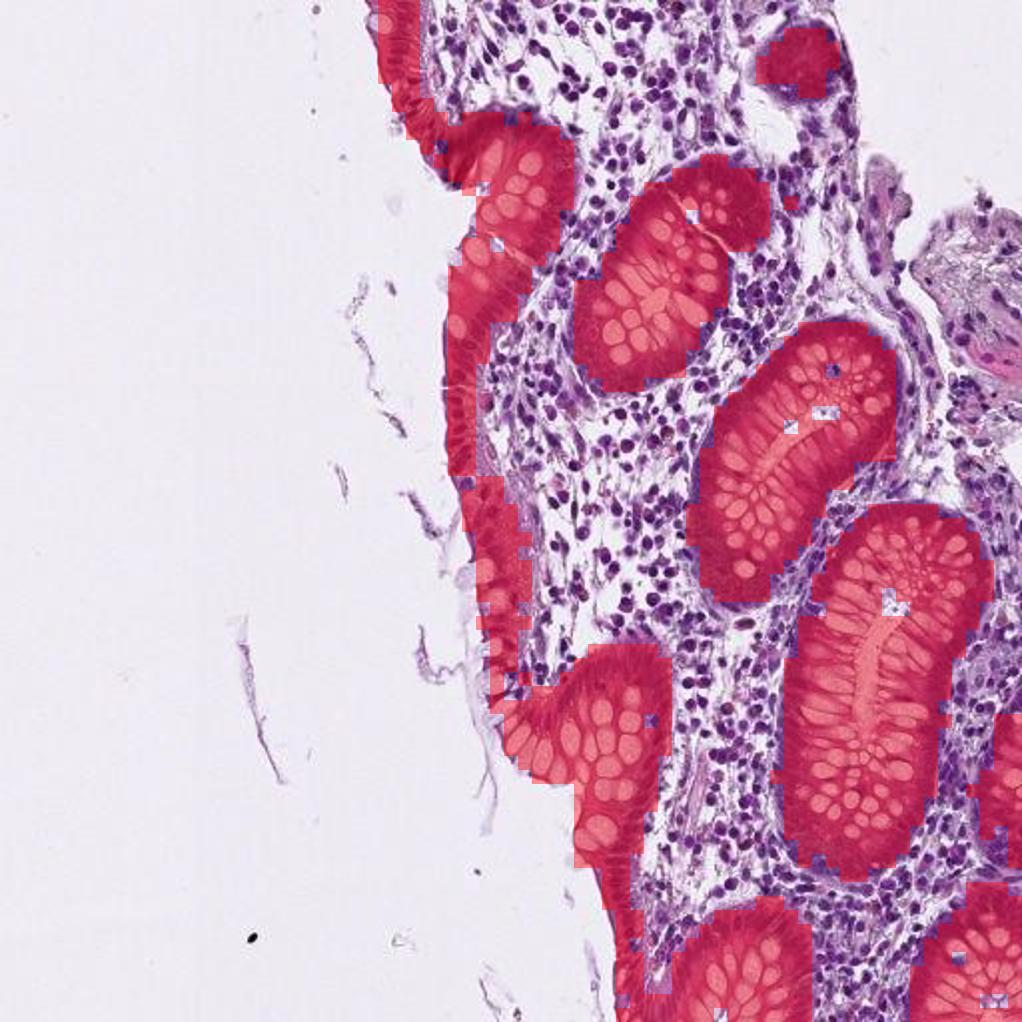}
        \end{minipage} &
        \begin{minipage}{0.33\textwidth}
            \centering
            \includegraphics[width=\linewidth]{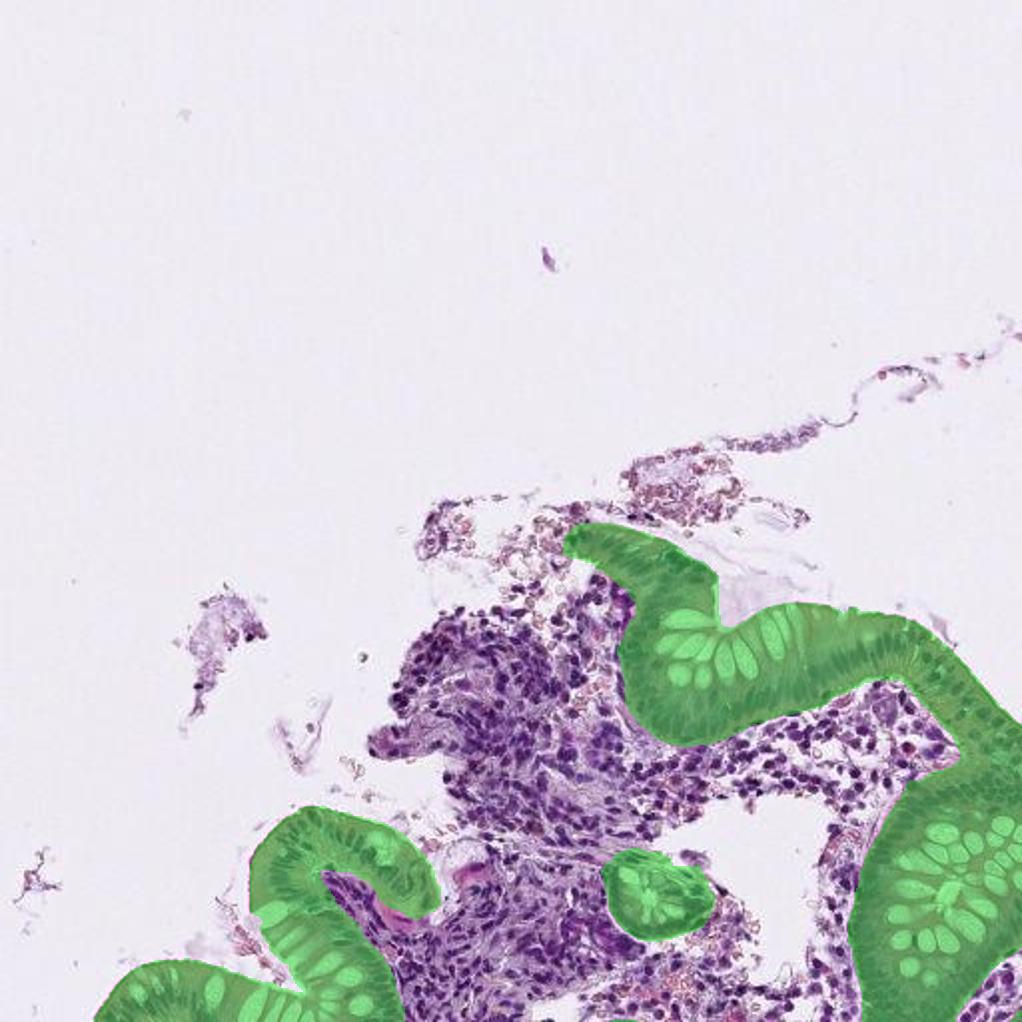}\\
            \includegraphics[width=\linewidth]{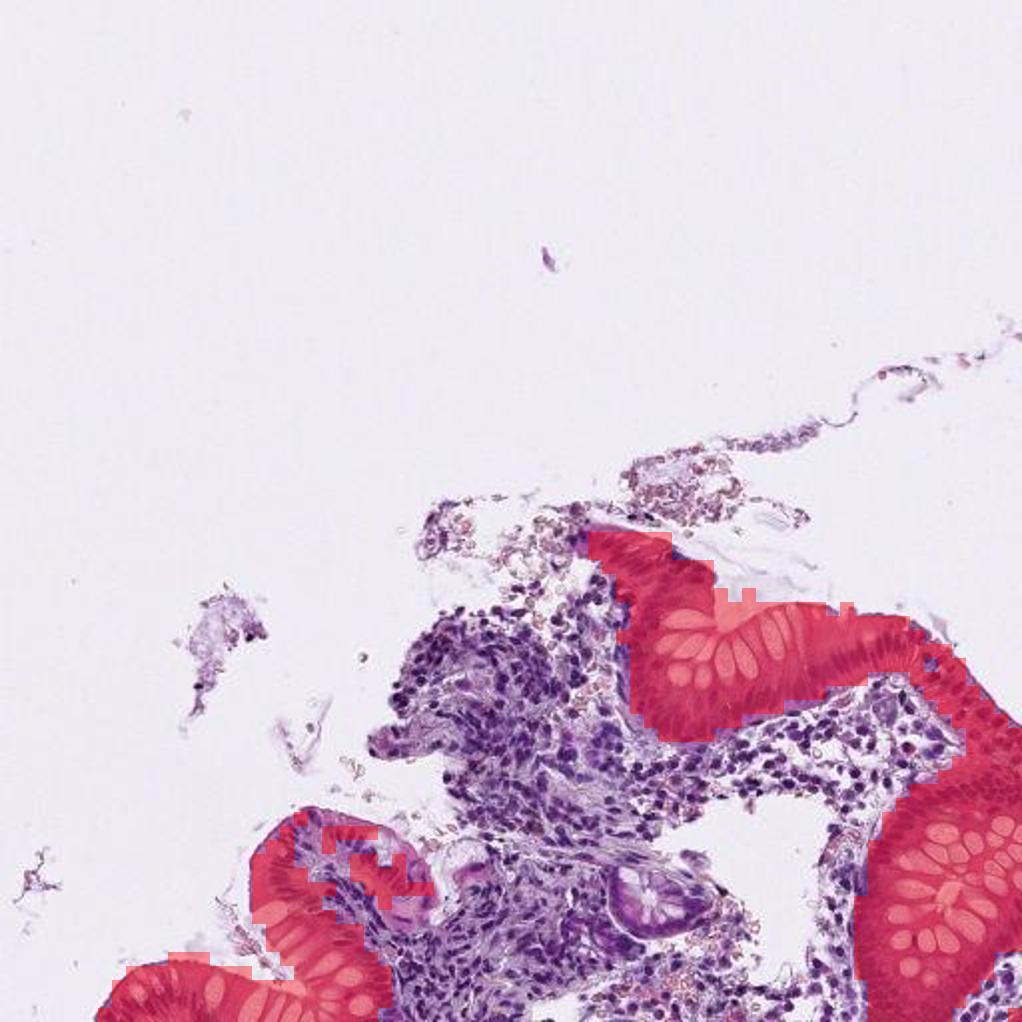}
        \end{minipage}&
        \begin{minipage}{0.33\textwidth}
            \centering
            \includegraphics[width=\linewidth]{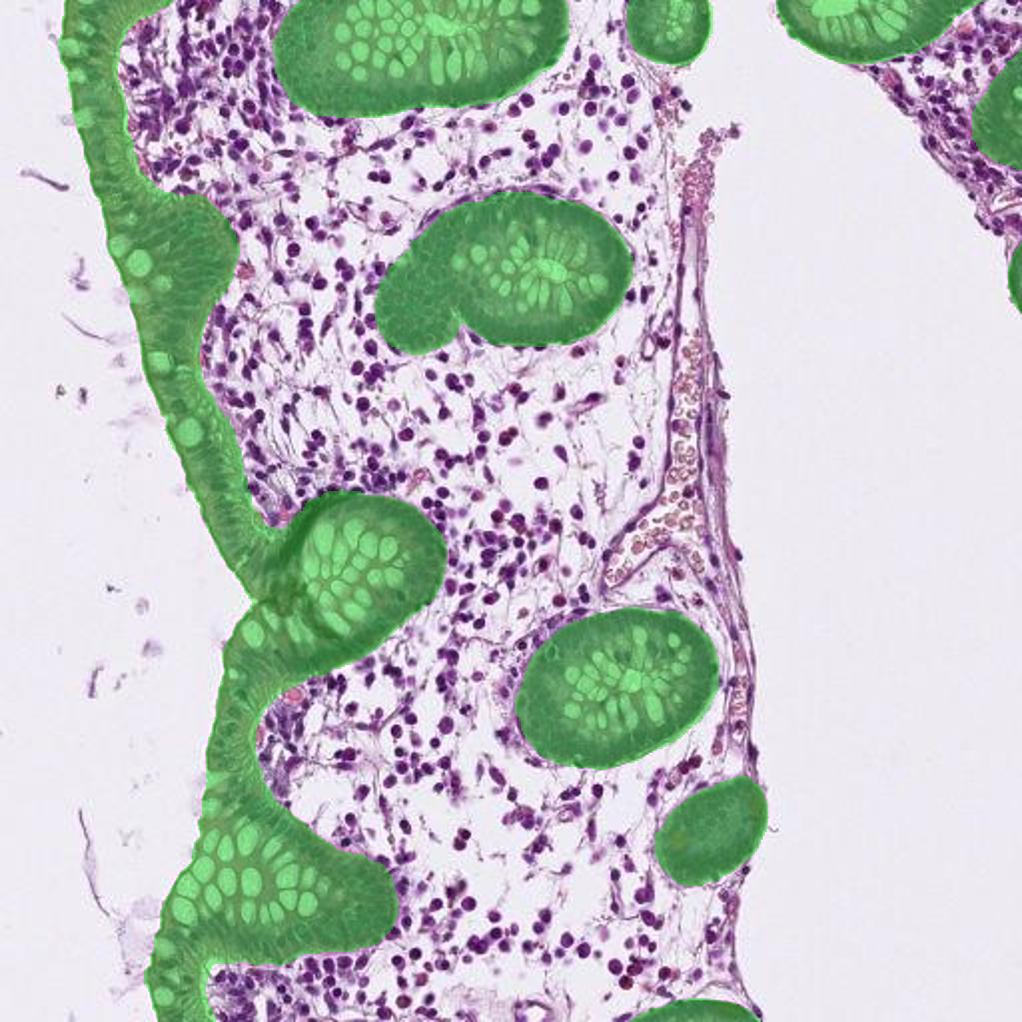}\\
            \includegraphics[width=\linewidth]{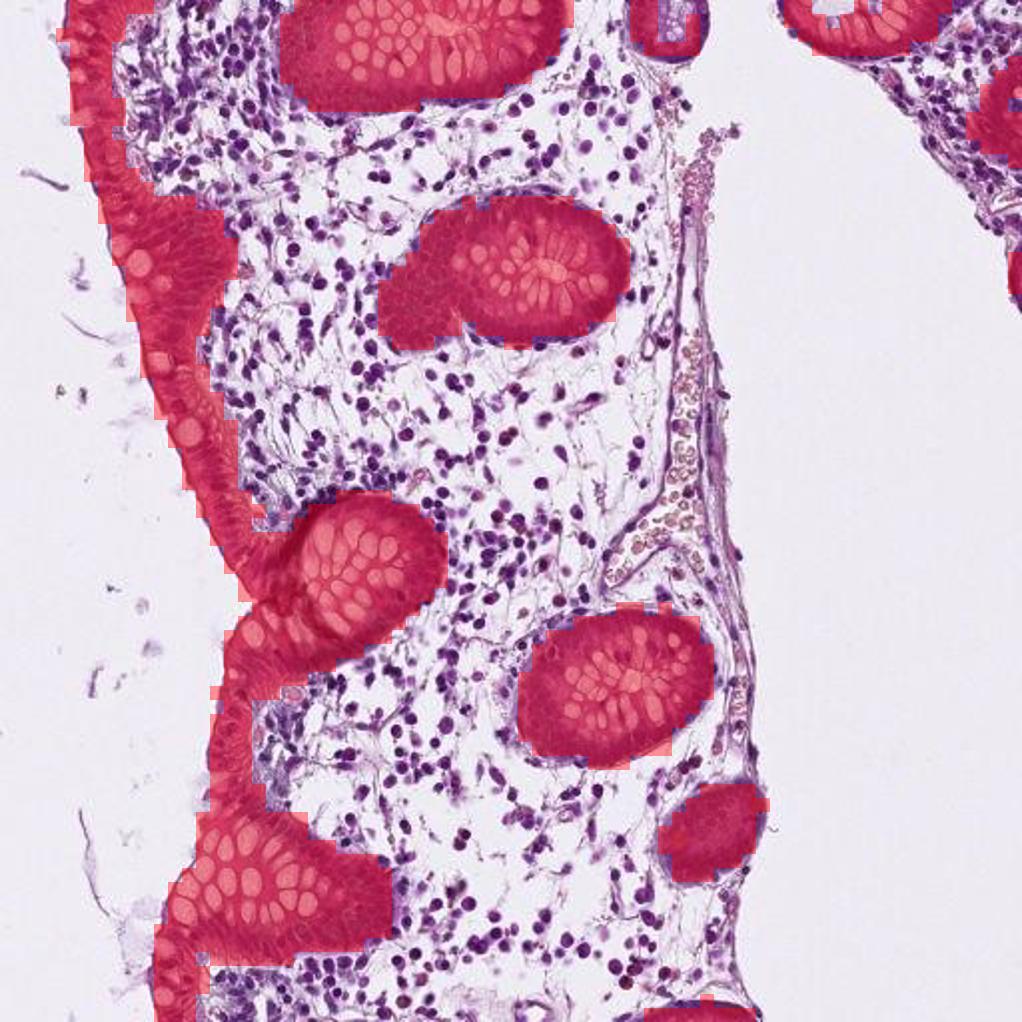}
        \end{minipage}
    \end{tabular}
  \caption{Visualizations of EpiSeg's predictions for the presence of epithelium at patch level (size $14 \times 14$ at a spatial resolution of 0.5 mpp). First row (in green), is the ground truth, and second row (in red) is the model's predictions.}
  \label{fig:episeg_gt_pred}

\end{figure}

%\begin{figure*}[t]
%    \centering
%    % Left column: Precision–Recall curve
%    \begin{minipage}[t]{0.38\textwidth}
%        \centering
%        \includegraphics[width=\linewidth]{figures/episeg_curve.png}
%        \caption*{Precision–Recall Curve}
%    \end{minipage}
%    \hfill
%    % Right column: 2×2 grid of GT/Pred images
%    \begin{minipage}[t]{0.58\textwidth}
%        \centering
%        \begin{tabular}{cc}
%            \begin{minipage}{0.48\textwidth}
%                \centering
%                \includegraphics[width=\linewidth]{figures/episeg_gt_1.png}\\
%                \includegraphics[width=\linewidth]{figures/episeg_pred_1.png}
%            \end{minipage} &
%            \begin{minipage}{0.48\textwidth}
%                \centering
%                \includegraphics[width=\linewidth]{figures/episeg_gt_2.png}\\
%                \includegraphics[width=\linewidth]{figures/episeg_pred_2.png}
%            \end{minipage}
%        \end{tabular}
%        \caption*{Ground Truth / Prediction Examples}
%    \end{minipage}
%\end{figure*}

\begin{figure}[ht!]
 % Caption and label go in the first argument and the figure contents
 % go in the second argument
    \begin{tabular}{c}
        \includegraphics[width=0.96\linewidth]{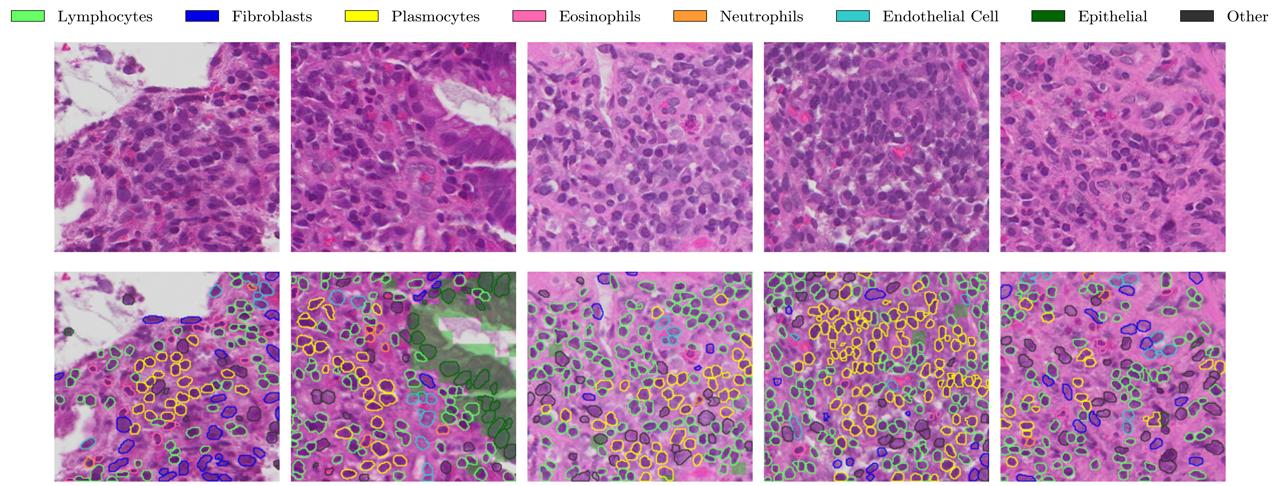} \\
        \includegraphics[width=0.96\linewidth]{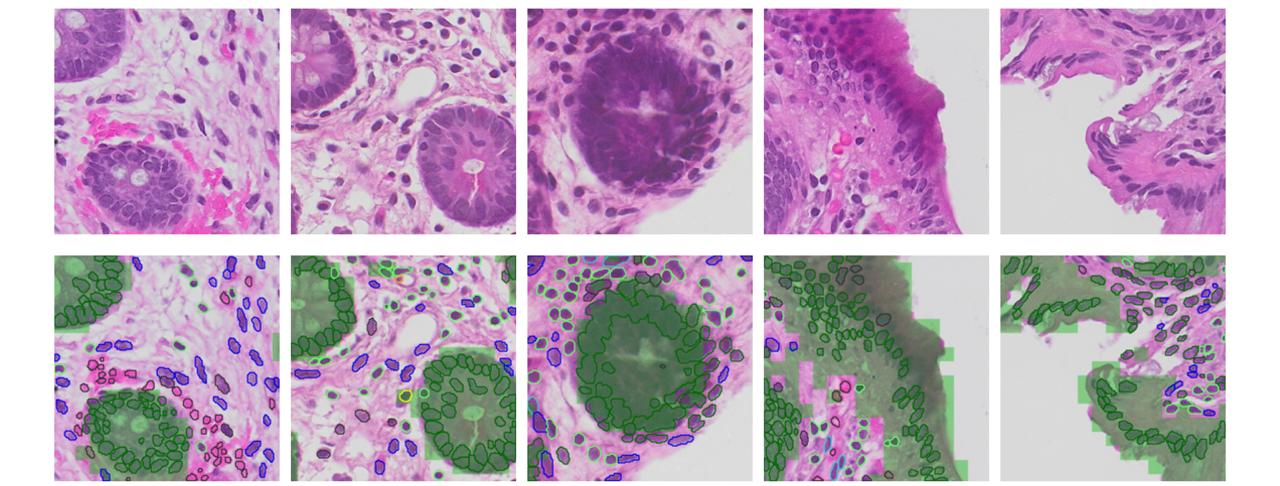}
    \end{tabular}
  \caption{Examples of max (top two rows) and min (bottom two rows) tiles from the SPARC IBD cohort, with overlays of the predictions from HistoPLUS (cell contours) and EpiSeg (epithelium zones in green).}
  \label{fig:min_max_ccf}
\end{figure}

\paragraph{The interpretability modules of IMILIA provide consistent tile-level patterns across cohorts.}
Tile scores were computed for all slides, allowing to extract max and min tiles for each dataset. Interpretable features were then computed for each extreme tile, including cell type counts (from HistoPLUS predictions) and density of several immune cell types in epithelium (through a combination of HistoPLUS and EpiSeg). Figure \ref{fig:min_max_ccf} illustrates extreme tiles sampled randomly within the min and max subgroups in the SPARC IBD cohort, with the predictions of HistoPLUS and EpiSeg. Additional visualization examples for the external validation cohorts FINBB and IBDColEPI can be found in Appendix \ref{app:min_max_val}.

\begin{figure}[ht!]
 % Caption and label go in the first argument and the figure contents
 % go in the second argument
    \begin{tabular}{ccc}
        \begin{minipage}{0.32\textwidth}
            \centering
            \includegraphics[width=\linewidth]{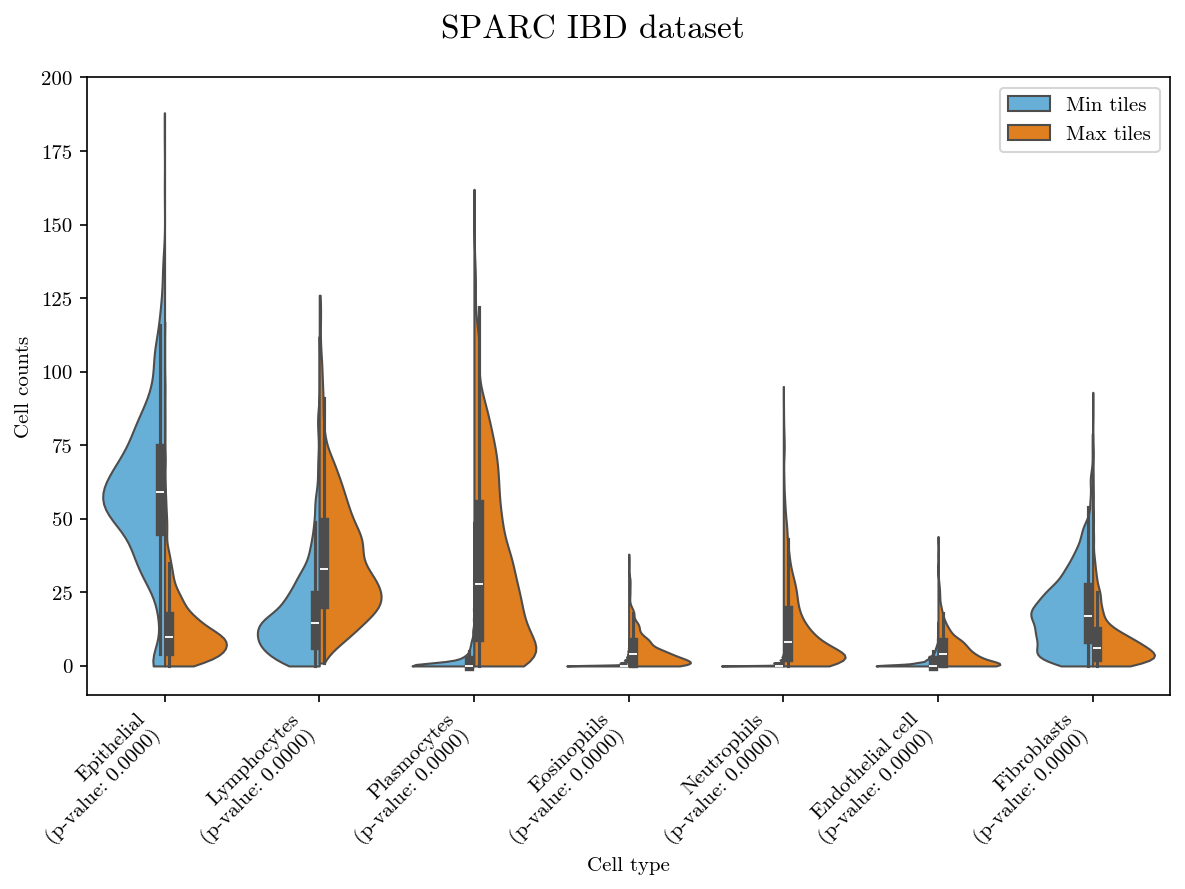}
        \end{minipage} &
        \begin{minipage}{0.32\textwidth}
            \centering
            \includegraphics[width=\linewidth]{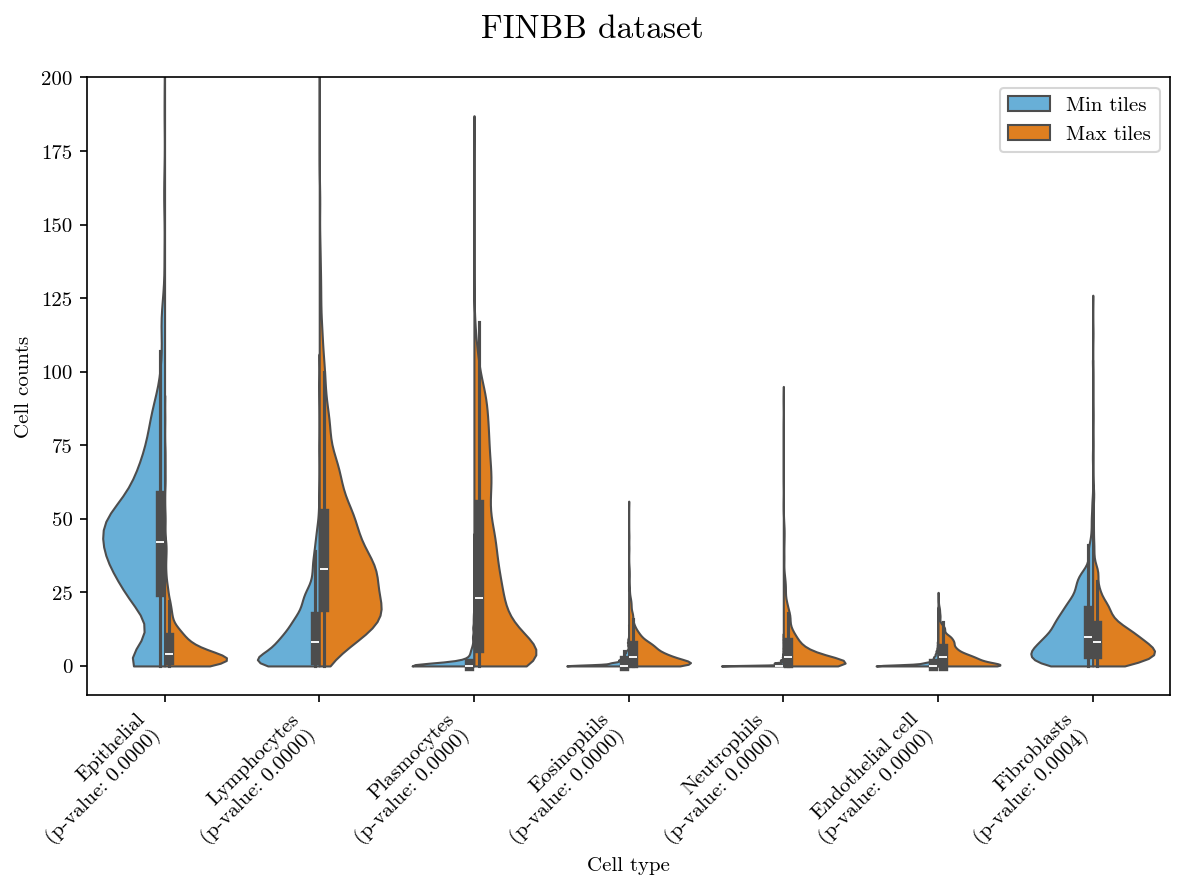}
        \end{minipage}&
        \begin{minipage}{0.32\textwidth}
            \centering
            \includegraphics[width=\linewidth]{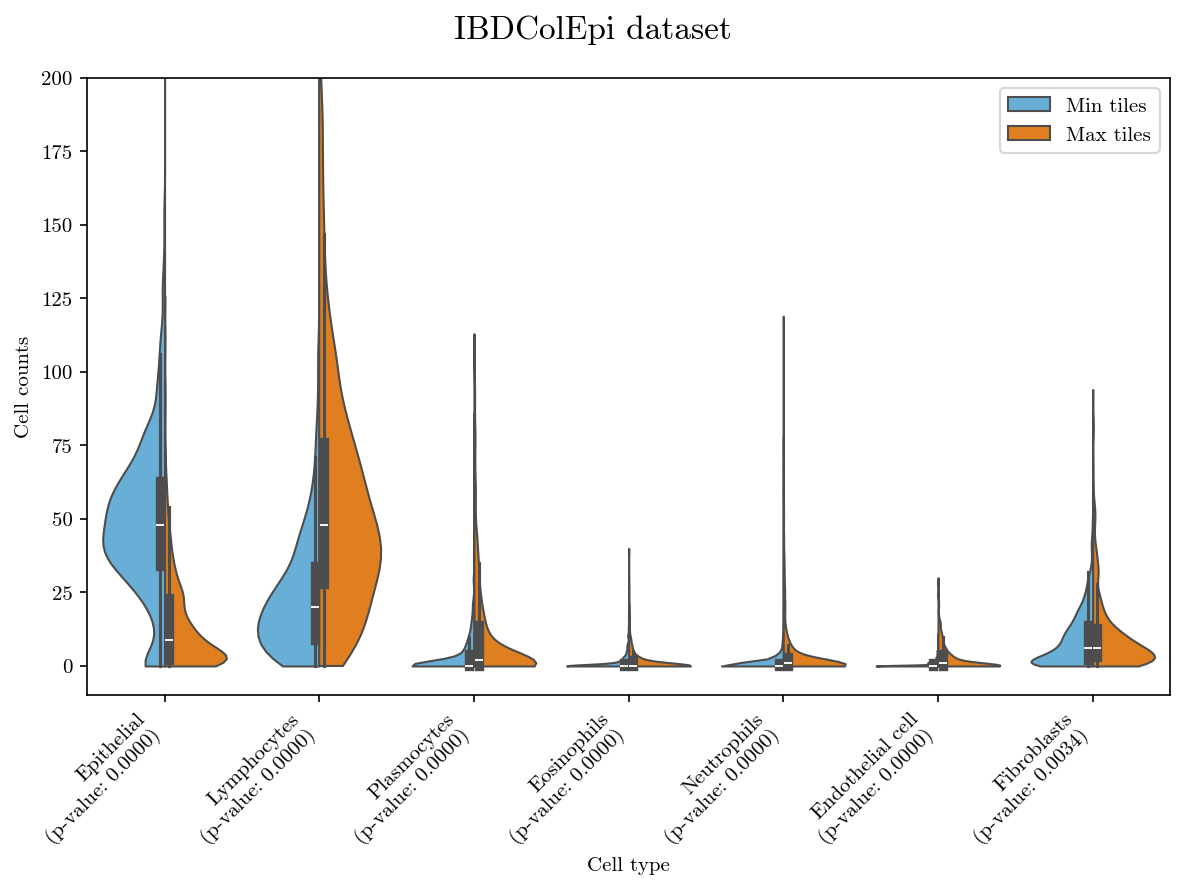}
        \end{minipage}
    \end{tabular}
  \caption{Violin plots of the distribution of cell types within min and max tiles, as analyzed by HistoPLUS, and by cohort.}
  \label{fig:min_max_comp}
\end{figure}

Figure \ref{fig:min_max_comp} shows the average composition of minimal and maximal tiles in terms of cell types counts. Minimal tiles are richer in epithelial cells, while maximal tiles have a higher concentration of immune cells (lymphocytes, plasmocytes, eosinophils, neutrophils) and endothelial cells. These patterns were consistent across all three datasets, indicating consistent interpretability of IMILIA. Visual examination of the minimal and maximal tile samples by an expert pathologist corroborated the quantitative findings derived from HistoPLUS predictions. To avoid bias, the pathologist was blinded to the model-predicted cell counts during the assessment. Minimal tiles predominantly represented non-inflamed tissue, characterized by epithelial areas with crypts and collagen-rich stroma (primarily dense collagen with occasional myxoid areas). In contrast, maximal tiles exhibited dense lymphocytic and plasmacytic inflammation, with scattered eosinophils and neutrophils and minimal epithelial tissue. Some maximal tiles also showed the presence of red blood cells and/or hemorrhage.

\begin{figure}[ht!]
 % Caption and label go in the first argument and the figure contents
 % go in the second argument
    \begin{tabular}{ccc}
        \begin{minipage}{0.32\textwidth}
            \centering
            \includegraphics[width=\linewidth]{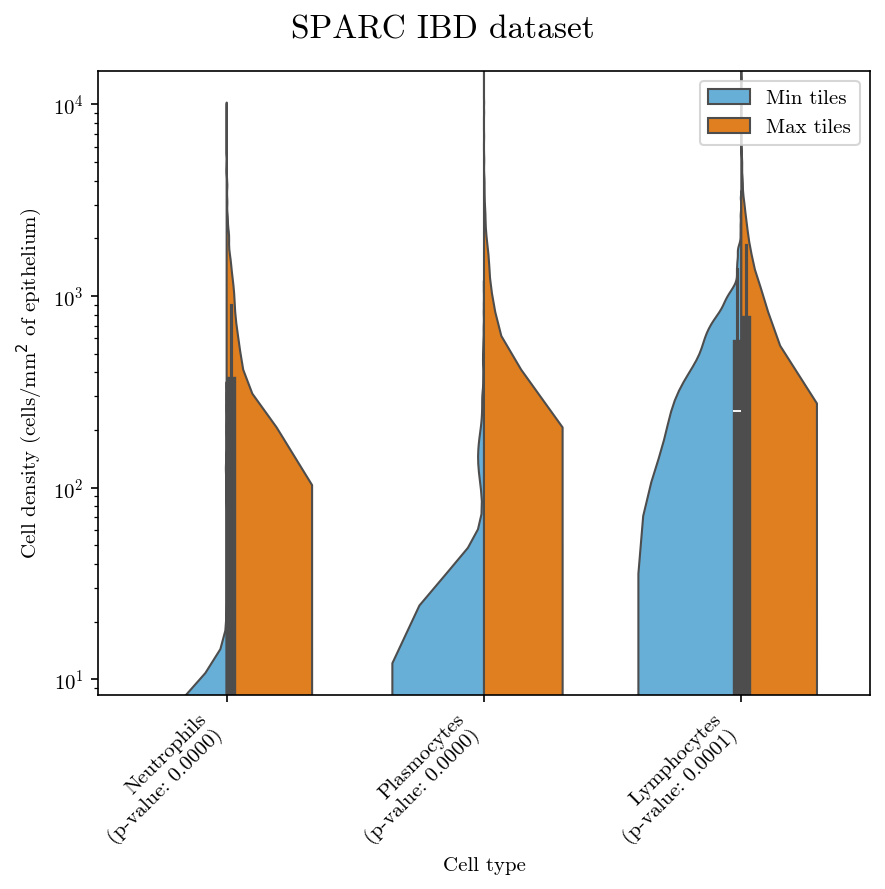}
        \end{minipage} &
        \begin{minipage}{0.32\textwidth}
            \centering
            \includegraphics[width=\linewidth]{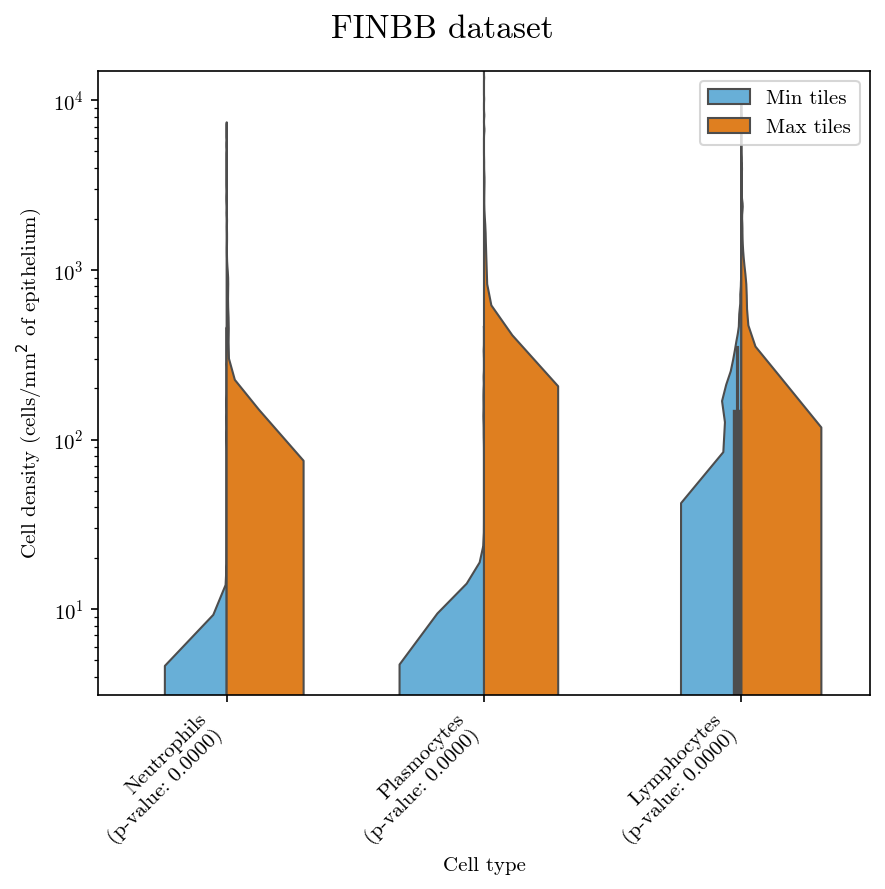}
        \end{minipage}&
        \begin{minipage}{0.32\textwidth}
            \centering
            \includegraphics[width=\linewidth]{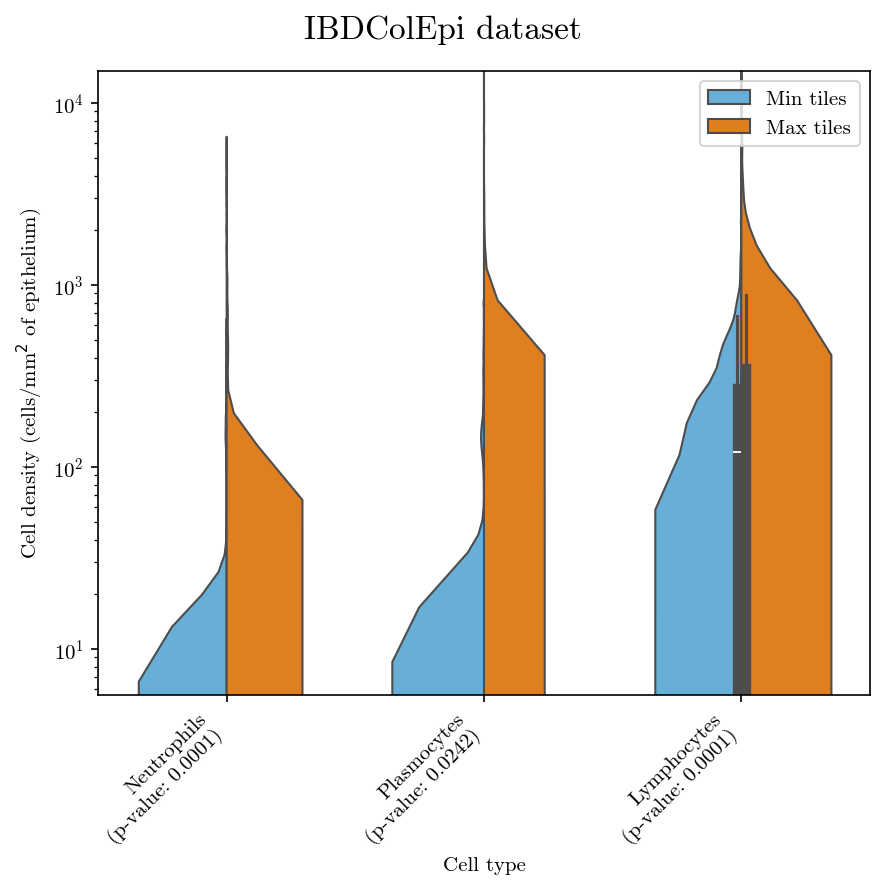}
        \end{minipage}
    \end{tabular}
  \caption{Violin plots of the distribution of immune cells in the epithelium.}
  \label{fig:neutrophils}
\end{figure}

The combination of HistoPLUS and EpiSeg allows to derive refined markers of the presence of immune cells in the epithelium, expressed as the number of cells per unit surface area of epithelium, integrating a location awareness dimension. Comparison between extreme tiles shows that the difference is most significant for the density of neutrophils in the epithelium, as illustrated in Figure \ref{fig:neutrophils}, and in line with known clinical indicators of the presence of inflammation in IBD.

% not working??
% \begin{figure}[ht!]
%  % Caption and label go in the first argument and the figure contents
%  % go in the second argument
% \floatconts
%   {fig:min_max_preds_ccf}
%   {\caption{Min (top) and max (bottom) tiles examples for the SPARC IBD dataset. Light green patch-wise overlay are the epithelium predictions of EpiSeg, and the cell types predicted by HistoPLUS are delineated in color.}}
%   {
%     % \includegraphics[width=0.95\linewidth]{figures/min_tiles_ccf.png}\\
%     \includegraphics[width=\linewidth]{figures/max_tiles_ccf.png}
%   }
% \end{figure}

\section{Conclusion}

In this study, we introduced IMILIA, an interpretable Multiple Instance Learning framework for the automated assessment of histological inflammation in IBD. Our results demonstrate that the pipeline not only achieves robust classification performance across diverse datasets but also, and crucially, provides transparent biological justifications for its predictions through quantitative cell and tissue-level analysis. One limitation of standard MIL approaches is the reliance on visual attention heatmaps for interpretability, which can lead to confirmation bias from the observer. IMILIA overcomes this by coupling the MIL attention mechanism with the HistoPlus and EpiSeg modules, automatically deriving interpretable markers for the presence of inflammation, a first step towards automated biomarker discovery in IBD.

We also note the following limitations of our study. First, the interpretability module, fully automated, depends on the intrinsic performance of its two blocks, HistoPLUS and EpiSeg. While their performance can be estimated, inaccurate predictions can be a source of uncertainty in the full model, which would require some further validation in view of clinical usability. Second, all our models relied on a foundation model (H0-mini) developed on oncology WSIs. While it shows the strong transferability of such FMs to a new therapeutic area, future research could help assess the need for FMs tailored to IBD or inflammatory conditions in general. Finally, and as a line of future research, we note that the label “inflamed” used in this study can hide various histological patterns (\emph{e.g.}, ulcers, architectural distorsion, immune infiltration, crypt abscess, granulomas, etc.), and extending the model to these various categories is a natural perspective.

\clearpage  % Acknowledgements, references, and appendix do not count toward the page limit (if any)
\begin{credits}
\subsubsection{\ackname}
% Acknowledgments---Will not appear in anonymized version
The results published here are in part from the Study of a Prospective Adult Research Cohort with IBD (SPARC IBD). SPARC IBD is a component of the Crohn’s \& Colitis Foundation’s IBD Plexus data exchange platform. SPARC IBD enrolls patients with an established or new diagnosis of IBD from sites throughout the United States and links data collected from the electronic health record and study specific case report forms. Patients also provide blood, stool and biopsy samples at selected times during follow-up. The design and implementation of the SPARC IBD cohort has been previously described in \cite{Sparcibd2021}. It is available upon approved application to the Crohn’s \& Colitis Foundation IBD Plexus.

This study utilized the national Fingenious service (\url{www.fingenious.fi}) managed by Finnish Biobank Cooperative – FINBB (\url{www.finbb.fi}). This dataset is part of a large-scale effort to advance precision medicine and research into various diseases, including Inflammatory Bowel Disease (IBD) and other conditions. The following biobanks are acknowledged for delivering samples and data to the study: Helsinki Biobank (\url{https://www.helsinginbiopankki.fi}) and Finnish Clinical Biobank Tampere (\url{https://www.tays.fi/en-US/Research_and_development/Finnish_Clinical_Biobank_Tampere}).
\end{credits}

\bibliographystyle{splncs04}
\bibliography{imilia_bib}

@misc{adjadj2025,
      title={Towards Comprehensive Cellular Characterisation of H\&E slides}, 
      author={Benjamin Adjadj and Pierre-Antoine Bannier and Guillaume Horent and Sebastien Mandela and Aurore Lyon and Kathryn Schutte and Ulysse Marteau and Valentin Gaury and Laura Dumont and Thomas Mathieu and MOSAIC consortium and Reda Belbahri and Benoît Schmauch and Eric Durand and Katharina Von Loga and Lucie Gillet},
      year={2025},
      eprint={2508.09926},
      archivePrefix={arXiv},
      primaryClass={cs.CV},
      url={https://arxiv.org/abs/2508.09926}, 
}

@article{angyal2025,
	article-number = {2485},
	author = {Angyal, Dorottya and Balogh, Fruzsina and Bessissow, Talat and Wetwittayakhlang, Panu and Ilias, Akos and Gonczi, Lorant and Lakatos, Peter L.},
	doi = {10.3390/jcm14072485},
	issn = {2077-0383},
	journal = {Journal of Clinical Medicine},
	number = {7},
	pubmedid = {40217934},
	title = {The Role of Histology Alongside Clinical and Endoscopic Evaluation in the Management of IBD---A Narrative Review},
	url = {https://www.mdpi.com/2077-0383/14/7/2485},
	volume = {14},
	year = {2025}}

@article{burisch2025,
    author = {Burisch, J and Claytor, J and Hernandez, I and Hou, JK and Kaplan GG.},
    title = {The Cost of Inflammatory Bowel Disease Care: How to Make it Sustainable.},
    journal = {Clin Gastroenterol Hepatol.},
    year = 2025,
    volume = {23(3)},
    pages = {386-395},
    doi = {doi:10.1016/j.cgh.2024.06.049},
}

@article{Chen2024,
author={Chen, Richard J.
and Ding, Tong
and Lu, Ming Y.
and Williamson, Drew F. K.
and Jaume, Guillaume
and Song, Andrew H.
and Chen, Bowen
and Zhang, Andrew
and Shao, Daniel
and Shaban, Muhammad
and Williams, Mane
and Oldenburg, Lukas
and Weishaupt, Luca L.
and Wang, Judy J.
and Vaidya, Anurag
and Le, Long Phi
and Gerber, Georg
and Sahai, Sharifa
and Williams, Walt
and Mahmood, Faisal},
title={Towards a general-purpose foundation model for computational pathology},
journal={Nature Medicine},
year={2024},
month={Mar},
day={01},
volume={30},
number={3},
pages={850-862},
issn={1546-170X},
doi={10.1038/s41591-024-02857-3},
url={https://doi.org/10.1038/s41591-024-02857-3}
}

@misc{courtiol2020,
      title={Classification and Disease Localization in Histopathology Using Only Global Labels: A Weakly-Supervised Approach}, 
      author={Pierre Courtiol and Eric W. Tramel and Marc Sanselme and Gilles Wainrib},
      year={2020},
      eprint={1802.02212},
      archivePrefix={arXiv},
      primaryClass={cs.CV},
      url={https://arxiv.org/abs/1802.02212}, 
}

@INPROCEEDINGS{Ding2025,
  author={Ding, Ruiwen and Li, Lin and Soans, Rajath and Shah, Tosha and Krishnan, Radha and Sze, Marc Alexander and Lukyanov, Sasha and Deshpande, Yash and Chen, Antong},
  booktitle={2025 IEEE 22nd International Symposium on Biomedical Imaging (ISBI)}, 
  title={Predicting Ulcer in H\&E Images of Inflammatory Bowel Disease Using Domain-Knowledge-Driven Graph Neural Network}, 
  year={2025},
  volume={},
  number={},
  pages={1-5},
  doi={10.1109/ISBI60581.2025.10980783}}

@article{Divicenzo2024,
    author = {Di Vincenzo, Federica and Quintero, Maria A and Serigado, Joao M and Koru-Sengul, Tulay and Killian, Rose Marie and Poveda, Julio and England, Jonathan and Damas, Oriana and Kerman, David and Deshpande, Amar and Abreu, Maria T},
    title = {Histologic and Endoscopic Findings Are Highly Correlated in a Prospective Cohort of Patients With Inflammatory Bowel Diseases},
    journal = {Journal of Crohn's and Colitis},
    volume = {19},
    number = {6},
    pages = {jjae141},
    year = {2024},
    month = {12},
    issn = {1876-4479},
    doi = {10.1093/ecco-jcc/jjae141},
    url = {https://doi.org/10.1093/ecco-jcc/jjae141},
    eprint = {https://academic.oup.com/ecco-jcc/article-pdf/19/6/jjae141/61302303/jjae141.pdf},
}

@article{Elta1987,
    author={Elta, Grace H. and Appelman, Henry D. and Behler, Elizabeth M. and Wilson, Joanne A. P. and Nostrant, Timothy J.},
    year = {1987},
    title = {A Study of the Correlation between Endoscopic and Histological Diagnoses in Gastroduodenitis},
    journal = {The American Journal of Gastroenterology},
    volume = {82},
    number = {8},
    pages = {749-753},
    url = {http://hdl.handle.net/2027.42/72433}
}

@article{Feakins2023,
    author = {Feakins, Roger and Borralho Nunes, Paula and Driessen, Ann and Gordon, Ilyssa O and Zidar, Nina and Baldin, Pamela and Christensen, Britt and Danese, Silvio and Herlihy, Naoimh and Iacucci, Marietta and Loughrey, Maurice B and Magro, Fernando and Mookhoek, Aart and Svrcek, Magali and Rosini, Francesca},
    title = {Definitions of Histological Abnormalities in Inflammatory Bowel Disease: an ECCO Position Paper},
    journal = {Journal of Crohn's and Colitis},
    volume = {18},
    number = {2},
    pages = {175-191},
    year = {2023},
    month = {08},
    issn = {1873-9946},
    doi = {10.1093/ecco-jcc/jjad142},
    url = {https://doi.org/10.1093/ecco-jcc/jjad142},
    eprint = {https://academic.oup.com/ecco-jcc/article-pdf/18/2/175/56759649/jjad142.pdf},
}

@Article{Feuerstein2014,
author={Feuerstein, Joseph D.
and Cheifetz, Adam S.},
title={Ulcerative Colitis: Epidemiology, Diagnosis, and Management},
journal={Mayo Clinic Proceedings},
year={2014},
month={Nov},
day={01},
publisher={Elsevier},
volume={89},
number={11},
pages={1553-1563},
issn={0025-6196},
doi={10.1016/j.mayocp.2014.07.002},
url={https://doi.org/10.1016/j.mayocp.2014.07.002}
}

@misc{filiot2025,
  title={Distilling foundation models for robust and efficient models in digital pathology}, 
  author={Alexandre Filiot and Nicolas Dop and Oussama Tchita and Auriane Riou and Rémy Dubois and Thomas Peeters and Daria Valter and Marin Scalbert and Charlie Saillard and Geneviève Robin and Antoine Olivier},
  year={2025},
  eprint={2501.16239},
  archivePrefix={arXiv},
  primaryClass={cs.CV},
  url={https://arxiv.org/abs/2501.16239}, 
}

@article{Geboes2000,
	author = {Geboes, K and Riddell, R and {\"O}st, A and Jensfelt, B and Persson, T and L{\"o}fberg, R},
	title = {A reproducible grading scale for histological assessment of inflammation in ulcerative colitis},
	volume = {47},
	number = {3},
	pages = {404--409},
	year = {2000},
	doi = {10.1136/gut.47.3.404},
	publisher = {BMJ Publishing Group},
	issn = {0017-5749},
	URL = {https://gut.bmj.com/content/47/3/404},
	eprint = {https://gut.bmj.com/content/47/3/404.full.pdf},
	journal = {Gut}
}

@Article{Graham2020,
author={Graham, Daniel B.
and Xavier, Ramnik J.},
title={Pathway paradigms revealed from the genetics of inflammatory bowel disease},
journal={Nature},
year={2020},
month={Feb},
day={01},
volume={578},
number={7796},
pages={527-539},
issn={1476-4687},
doi={10.1038/s41586-020-2025-2},
url={https://doi.org/10.1038/s41586-020-2025-2}
}

@article{Horst2024,
title = {CellViT: Vision Transformers for precise cell segmentation and classification},
journal = {Medical Image Analysis},
volume = {94},
pages = {103143},
year = {2024},
issn = {1361-8415},
doi = {https://doi.org/10.1016/j.media.2024.103143},
url = {https://www.sciencedirect.com/science/article/pii/S1361841524000689},
author = {Fabian Hörst and Moritz Rempe and Lukas Heine and Constantin Seibold and Julius Keyl and Giulia Baldini and Selma Ugurel and Jens Siveke and Barbara Grünwald and Jan Egger and Jens Kleesiek},
}

@article{IACUCCI2023,
title = {Artificial Intelligence Enabled Histological Prediction of Remission or Activity and Clinical Outcomes in Ulcerative Colitis},
journal = {Gastroenterology},
volume = {164},
number = {7},
pages = {1180-1188.e2},
year = {2023},
issn = {0016-5085},
doi = {https://doi.org/10.1053/j.gastro.2023.02.031},
url = {https://www.sciencedirect.com/science/article/pii/S0016508523002160},
author = {Marietta Iacucci and Tommaso Lorenzo Parigi and Rocio {Del Amor} and Pablo Meseguer and Giulio Mandelli and Anna Bozzola and Alina Bazarova and Pradeep Bhandari and Raf Bisschops and Silvio Danese and Gert {De Hertogh} and Jose G. Ferraz and Martin Goetz and Enrico Grisan and Xianyong Gui and Bu Hayee and Ralf Kiesslich and Mark Lazarev and Remo Panaccione and Adolfo Parra-Blanco and Luca Pastorelli and Timo Rath and Elin S. Røyset and Gian Eugenio Tontini and Michael Vieth and Davide Zardo and Subrata Ghosh and Valery Naranjo and Vincenzo Villanacci},
}

@article{IACUCCI2021,
title = {An International Multicenter Real-Life Prospective Study of Electronic Chromoendoscopy Score PICaSSO in Ulcerative Colitis},
journal = {Gastroenterology},
volume = {160},
number = {5},
pages = {1558-1569.e8},
year = {2021},
issn = {0016-5085},
doi = {https://doi.org/10.1053/j.gastro.2020.12.024},
url = {https://www.sciencedirect.com/science/article/pii/S0016508520355633},
author = {Marietta Iacucci and Samuel C.L. Smith and Alina Bazarova and Uday N. Shivaji and Pradeep Bhandari and Rosanna Cannatelli and Marco Daperno and Jose Ferraz and Martin Goetz and Xianyong Gui and Bu Hayee and Gert {De Hertogh} and Mark Lazarev and Jim Li and Olga M. Nardone and Adolfo Parra-Blanco and Luca Pastorelli and Remo Panaccione and Vincenzo Occhipinti and Timo Rath and Gian Eugenio Tontini and Michael Vieth and Vincenzo Villanacci and Davide Zardo and Raf Bisschops and Ralf Kiesslich and Subrata Ghosh},
}

@InProceedings{ilse2018,
  title = 	 {Attention-based Deep Multiple Instance Learning},
  author =       {Ilse, Maximilian and Tomczak, Jakub and Welling, Max},
  booktitle = 	 {Proceedings of the 35th International Conference on Machine Learning},
  pages = 	 {2127--2136},
  year = 	 {2018},
  editor = 	 {Dy, Jennifer and Krause, Andreas},
  volume = 	 {80},
  series = 	 {Proceedings of Machine Learning Research},
  month = 	 {10--15 Jul},
  publisher =    {PMLR},
  url = 	 {https://proceedings.mlr.press/v80/ilse18a.html}
}

@article{Marchal-Bressenot2017,
	author = {Marchal-Bressenot, Aude and Salleron, Julia and Boulagnon-Rombi, Camille and Bastien, Claire and Cahn, Virginie and Cadiot, Guillaume and Diebold, Marie-Dani{\`e}le and Danese, Silvio and Reinisch, Walter and Schreiber, Stefan and Travis, Simon and Peyrin-Biroulet, Laurent},
	title = {Development and validation of the Nancy histological index for UC},
	volume = {66},
	number = {1},
	pages = {43--49},
	year = {2017},
	doi = {10.1136/gutjnl-2015-310187},
	publisher = {BMJ Publishing Group},
	issn = {0017-5749},
	URL = {https://gut.bmj.com/content/66/1/43},
	eprint = {https://gut.bmj.com/content/66/1/43.full.pdf},
	journal = {Gut}
}

@InProceedings{mokhtari24a,
  title = 	 {Interpretable histopathology-based prediction of disease relevant features in Inflammatory Bowel Disease biopsies using weakly-supervised deep learning},
  author =       {Mokhtari, Ricardo and Hamidinekoo, Azam and Sutton, Daniel James and Lewis, Arthur and Angermann, Bastian and Gehrmann, Ulf and Lundin, P\r{a}l and Adissu, Hibret and Cairns, Junmei and Neisen, Jessica and Khan, Emon and Marks, Daniel and Khachapuridze, Nia and Qaiser, Talha and Burlutskiy, Nikolay},
  booktitle = 	 {Medical Imaging with Deep Learning},
  pages = 	 {479--495},
  year = 	 {2024},
  editor = 	 {Oguz, Ipek and Noble, Jack and Li, Xiaoxiao and Styner, Martin and Baumgartner, Christian and Rusu, Mirabela and Heinmann, Tobias and Kontos, Despina and Landman, Bennett and Dawant, Benoit},
  volume = 	 {227},
  series = 	 {Proceedings of Machine Learning Research},
  month = 	 {10--12 Jul},
  publisher =    {PMLR},
  url = 	 {https://proceedings.mlr.press/v227/mokhtari24a.html}
}

@Article{Molodecky2012,
author={Molodecky, Natalie A.
and Soon, Ing Shian
and Rabi, Doreen M.
and Ghali, William A.
and Ferris, Mollie
and Chernoff, Greg
and Benchimol, Eric I.
and Panaccione, Remo
and Ghosh, Subrata
and Barkema, Herman W.
and Kaplan, Gilaad G.},
title={Increasing Incidence and Prevalence of the Inflammatory Bowel Diseases With Time, Based on Systematic Review},
journal={Gastroenterology},
year={2012},
month={Jan},
day={01},
publisher={Elsevier},
volume={142},
number={1},
pages={46-54.e42},
issn={0016-5085},
doi={10.1053/j.gastro.2011.10.001},
url={https://doi.org/10.1053/j.gastro.2011.10.001}
}

@Article{Ng2017,
author={Ng, Siew C.
and Shi, Hai Yun
and Hamidi, Nima
and Underwood, Fox E.
and Tang, Whitney
and Benchimol, Eric I.
and Panaccione, Remo
and Ghosh, Subrata
and Wu, Justin C. Y.
and Chan, Francis K. L.
and Sung, Joseph J. Y.
and Kaplan, Gilaad G.},
title={Worldwide incidence and prevalence of inflammatory bowel disease in the 21st century: a systematic review of population-based studies},
journal={The Lancet},
year={2017},
month={Dec},
day={23},
publisher={Elsevier},
volume={390},
number={10114},
pages={2769-2778},
issn={0140-6736},
doi={10.1016/S0140-6736(17)32448-0},
url={https://doi.org/10.1016/S0140-6736(17)32448-0}
}

@ARTICLE{Pettersen2022,
AUTHOR={Pettersen, Henrik Sahlin  and Belevich, Ilya  and Røyset, Elin Synnøve  and Smistad, Erik  and Simpson, Melanie Rae  and Jokitalo, Eija  and Reinertsen, Ingerid  and Bakke, Ingunn  and Pedersen, André },
TITLE={Code-Free Development and Deployment of Deep Segmentation Models for Digital Pathology},
JOURNAL={Frontiers in Medicine},
VOLUME={9},
YEAR={2022},
URL={https://www.frontiersin.org/journals/medicine/articles/10.3389/fmed.2021.816281},
DOI={10.3389/fmed.2021.816281},
ISSN={2296-858X},
}

@article{Peyrin-Biroulet2024,
author = {Peyrin-Biroulet, Laurent and Adsul, Shashi and Stancati, Andrea and Dehmeshki, Jamshid and Kubassova, Olga},
title = {An artificial intelligence-driven scoring system to measure histological disease activity in ulcerative colitis},
journal = {United European Gastroenterology Journal},
volume = {12},
number = {8},
pages = {1028-1033},
doi = {https://doi.org/10.1002/ueg2.12562},
url = {https://onlinelibrary.wiley.com/doi/abs/10.1002/ueg2.12562},
eprint = {https://onlinelibrary.wiley.com/doi/pdf/10.1002/ueg2.12562},
year = {2024}
}

@article{Sparcibd2021,
    author = {Raffals, Laura E and Saha, Sumona and Bewtra, Meenakshi and Norris, Cecile and Dobes, Angela and Heller, Caren and O’Charoen, Sirimon and Fehlmann, Tara and Sweeney, Sara and Weaver, Alandra and Bishu, Shrinivas and Cross, Raymond and Dassopoulos, Themistocles and Fischer, Monika and Yarur, Andres and Hudesman, David and Parakkal, Deepak and Duerr, Richard and Caldera, Freddy and Korzenik, Joshua and Pekow, Joel and Wells, Katerina and Bohm, Matthew and Perera, Lilani and Kaur, Manreet and Ciorba, Matthew and Snapper, Scott and Scoville, Elizabeth A and Dalal, Sushila and Wong, Uni and Lewis, James D},
    title = {The Development and Initial Findings of A Study of a Prospective Adult Research Cohort with Inflammatory Bowel Disease (SPARC IBD)},
    journal = {Inflammatory Bowel Diseases},
    volume = {28},
    number = {2},
    pages = {192-199},
    year = {2021},
    month = {08},
    issn = {1078-0998},
    doi = {10.1093/ibd/izab071},
    url = {https://doi.org/10.1093/ibd/izab071},
    eprint = {https://academic.oup.com/ibdjournal/article-pdf/28/2/192/52624528/izab071.pdf},
}

@misc{Reigle2024,
	author = {Reigle, James and Lopez-Nunez, Oscar and Drysdale, Erik and Abuquteish, Dua and Liu, Xiaoxuan and Putra, Juan and Erdman, Lauren and Griffiths, Anne M. and Prasath, Surya and Siddiqui, Iram and Dhaliwal, Jasbir},
	title = {Using Deep Learning to Automate Eosinophil Counting in Pediatric Ulcerative Colitis Histopathological Images},
	elocation-id = {2024.04.03.24305251},
	year = {2024},
	doi = {10.1101/2024.04.03.24305251},
	publisher = {Cold Spring Harbor Laboratory Press},
	URL = {https://www.medrxiv.org/content/early/2024/04/05/2024.04.03.24305251},
	eprint = {https://www.medrxiv.org/content/early/2024/04/05/2024.04.03.24305251.full.pdf},
	journal = {medRxiv}
}

@article{Rubin2024,
    author = {Rubin, David T and Kubassova, Olga and Weber, Christopher R and Adsul, Shashi and Freire, Marcelo and Biedermann, Luc and Koelzer, Viktor H and Bressler, Brian and Xiong, Wei and Niess, Jan H and Matter, Matthias S and Kopylov, Uri and Barshack, Iris and Mayer, Chen and Magro, Fernando and Carneiro, Fatima and Maharshak, Nitsan and Greenberg, Ariel and Hart, Simon and Dehmeshki, Jamshid and Peyrin-Biroulet, Laurent},
    title = {Deployment of an Artificial Intelligence Histology Tool to Aid Qualitative Assessment of Histopathology Using the Nancy Histopathology Index in Ulcerative Colitis},
    journal = {Inflammatory Bowel Diseases},
    volume = {31},
    number = {6},
    pages = {1630-1636},
    year = {2024},
    month = {09},
    issn = {1536-4844},
    doi = {10.1093/ibd/izae204},
    url = {https://doi.org/10.1093/ibd/izae204},
    eprint = {https://academic.oup.com/ibdjournal/article-pdf/31/6/1630/59144200/izae204.pdf},
}

@article{Rymarczyk2023,
    author = {Rymarczyk, Dawid and Schultz, Weiwei and Borowa, Adriana and Friedman, Joshua R and Danel, Tomasz and Branigan, Patrick and Chałupczak, Michał and Bracha, Anna and Krawiec, Tomasz and Warchoł, Michał and Li, Katherine and De Hertogh, Gert and Zieliński, Bartosz and Ghanem, Louis R and Stojmirovic, Aleksandar},
    title = {Deep Learning Models Capture Histological Disease Activity in Crohn’s Disease and Ulcerative Colitis with High Fidelity},
    journal = {Journal of Crohn's and Colitis},
    volume = {18},
    number = {4},
    pages = {604-614},
    year = {2023},
    month = {10},
    issn = {1873-9946},
    doi = {10.1093/ecco-jcc/jjad171},
    url = {https://doi.org/10.1093/ecco-jcc/jjad171},
    eprint = {https://academic.oup.com/ecco-jcc/article-pdf/18/4/604/57309515/jjad171.pdf},
}

@misc{Plattner2025,
	author = {Plattner, Christina and Sturm, Gregor and K{\"u}hl, Anja A. and Atreya, Raja and Carollo, Sandro and Gronauer, Raphael and Rieder, Dietmar and G{\"u}nther, Michael and Ormanns, Steffen and Manzl, Claudia and Wirtz, Stefan and Meneghetti, Asier Rabasco and Hegazy, Ahmed N. and Patankar, Jay V. and Carrero, Zunamys I. and TRR241 IBDome Consortium and Neurath, Markus F. and Kather, Jakob Nikolas and Becker, Christoph and Siegmund, Britta and Trajanoski, Zlatko},
	title = {IBDome: An integrated molecular, histopathological, and clinical atlas of inflammatory bowel diseases},
	elocation-id = {2025.03.26.645544},
	year = {2025},
	doi = {10.1101/2025.03.26.645544},
	publisher = {Cold Spring Harbor Laboratory},
	URL = {https://www.biorxiv.org/content/early/2025/04/10/2025.03.26.645544},
	eprint = {https://www.biorxiv.org/content/early/2025/04/10/2025.03.26.645544.full.pdf},
	journal = {bioRxiv}
}

@article{Villanacci2025,
title = {Histological healing in IBD: Ready for prime time?},
journal = {Digestive and Liver Disease},
volume = {57},
number = {5},
pages = {954-960},
year = {2025},
issn = {1590-8658},
doi = {https://doi.org/10.1016/j.dld.2025.01.039},
url = {https://www.sciencedirect.com/science/article/pii/S1590865825000404},
author = {Vincenzo Villanacci and Rachele {Del Sordo} and Sara Mino and Giorgia Locci and Gabrio Bassotti},
}

\appendix
\renewcommand{\theHsection}{Appendix.\arabic{section}}
\section{Hyperparameters}
\label{app:hyperparameters}

We report in Table \ref{tab:hyperparameters} a more complete list of hyperparameters used to train the various models.

\begin{table}[htbp]%
  \centering    
  \caption{Hyperparameters used to train the various models.}
  \label{tab:hyperparameters}
  \fontsize{10}{\baselineskip}\selectfont
    \begin{tabular}{cccc}
    \toprule
    \multicolumn{2}{c}{\textbf{Chowder model}} & \multicolumn{2}{c}{\textbf{EpiSeg}}\\
    \midrule
    Tile size & $224 \times 224$   & Tile size  & $1022 \times 1022$\\
    Batch size & 256 & Patch size & 14 \\
    Number of folds & 5 & Number of folds & 3 \\
    Number of channels ($K$) & 5 & $C$  & $10^{-2}$\\
    $r$ & 25 &  & \\
    Learning rate & 0.01 & & \\
    Max number of tiles & 1000 & & \\
    MLP hidden layers' neurons & [128, 64] & & \\
    MLP hidden layers' dropout & [0.5, 0.5] & & \\
    \bottomrule
    \end{tabular}
   
\end{table}

\section{Detailed HistoPLUS transfer performance}
\label{app:histoplus}

We report in Table \ref{tab:histoplus-detection-perf} detailed segmentation metrics for HistoPLUS, comparing its performance on the oncology validation dataset and on the SPARC IBD dataset. We refer to the original HistoPLUS paper for a precise definition of the metrics.
\begin{table}[htbp]
  \centering
  \caption{Performance of HistoPLUS in cell detection and segmentation for HistoVAL (oncology) and SPARC IBD. We report mean values and confidence intervals at 95\% level, obtained by bootstrapping with 1000 repeats.}
  \label{tab:histoplus-detection-perf}%
  \fontsize{9}{\baselineskip}\selectfont
   \begin{tabular}{lccc}
   \toprule
   \bfseries Dataset & \bfseries Panoptic Quality & \bfseries Detection Quality & \bfseries Segmentation Quality\\
   \midrule
   HistoVAL & 0.605 (0.595; 0.613) & 0.753 (0.742; 0.763) & 0.801 (0.799; 0.804)\\
   SPARC IBD & 0.586 (0.572; 0.599) & 0.774 (0.760; 0.789) & 0.755 (0.749; 0.761) \\
   \bottomrule
   \end{tabular}
\end{table}

\section{EpiSeg precision-recall curve}
\label{app:pr-curve}

\begin{figure}[ht!]
 % Caption and label go in the first argument and the figure contents
 % go in the second argument
    \centering
    \begin{tabular}{c}
        \includegraphics[width=0.7\linewidth]{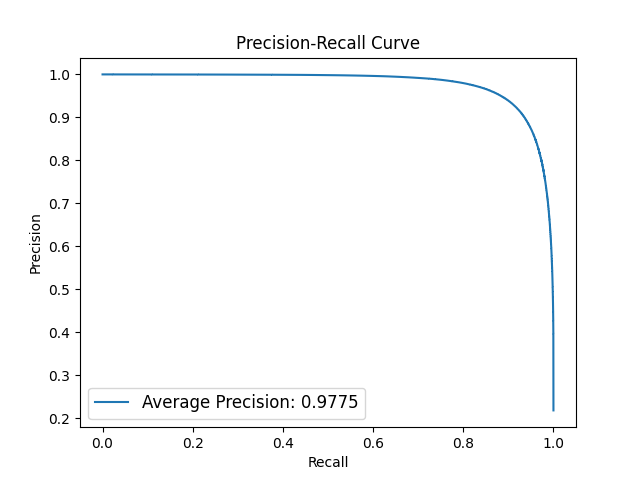} \\
    \end{tabular}
  \caption{Precision-recall curve for the patch-level prediction of presence of epithelium by EpiSeg on the IBDColEpi test set. The average precision metric is defined as the area under the precision-recall curve.}
  \label{fig:pr-curve}
\end{figure}

Figure \ref{fig:pr-curve} shows the Precision-Recall curve for the EpiSeg model on the IBDColEpi test set.

\section{Min and Max tiles visualization on the external validation cohorts}
\label{app:min_max_val}

\begin{figure}[ht!]
 % Caption and label go in the first argument and the figure contents
 % go in the second argument
    \begin{tabular}{c}
        \includegraphics[width=0.96\linewidth]{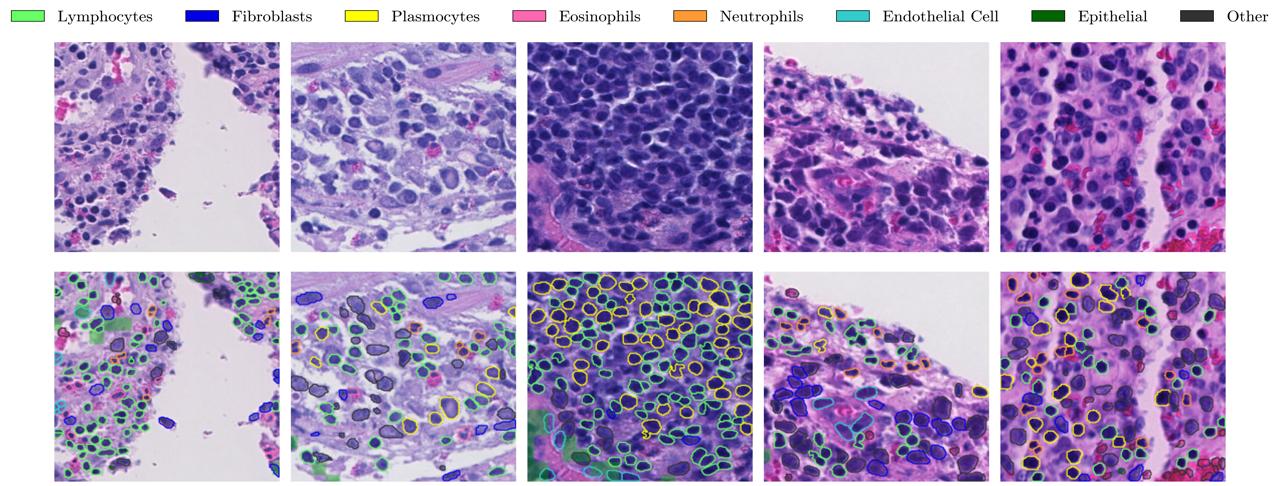} \\
        \includegraphics[width=0.96\linewidth]{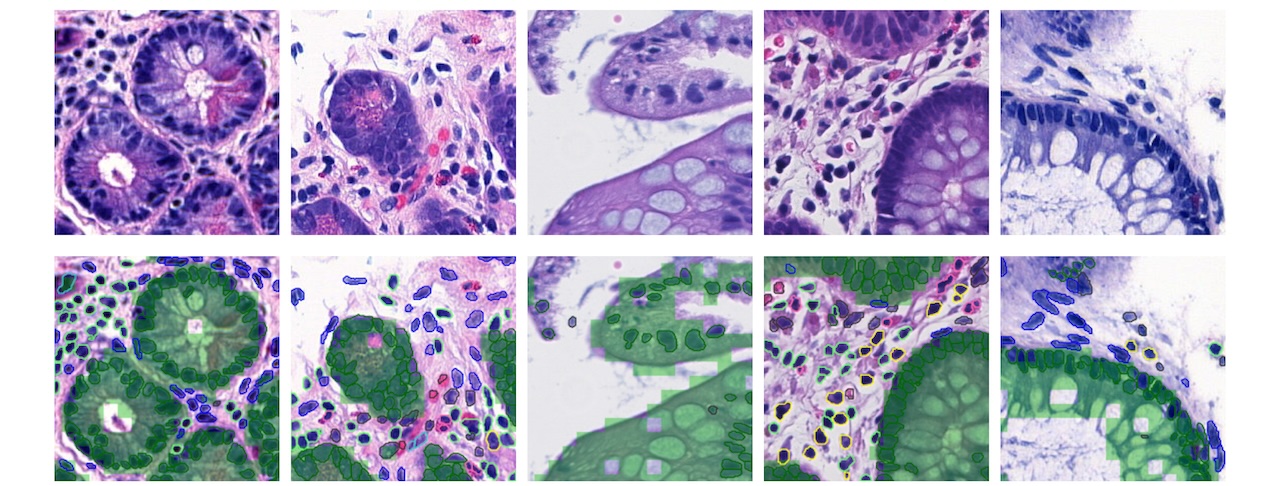}
    \end{tabular}
  \caption{Max (top two rows) and min (bottom two rows) tiles within the FINBB cohort, with overlays of the predictions from HistoPLUS and EpiSeg. Interestingly, one can notice the staining distribution shift compared to discovery cohort.}
  \label{fig:min_max_finbb}
\end{figure}

\begin{figure}[ht!]
 % Caption and label go in the first argument and the figure contents
 % go in the second argument
    \begin{tabular}{c}
        \includegraphics[width=0.96\linewidth]{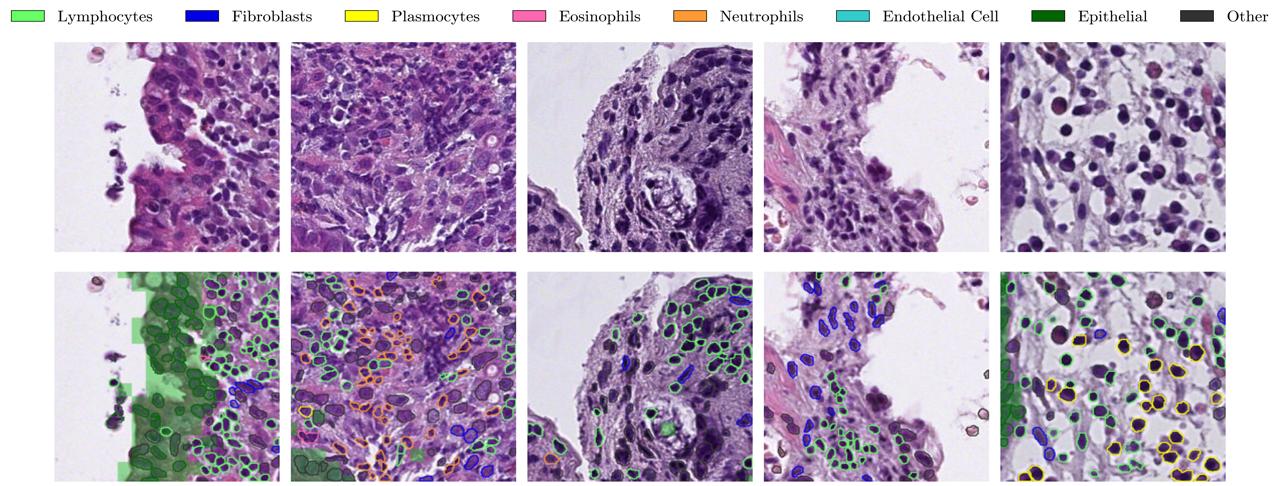} \\
        \includegraphics[width=0.96\linewidth]{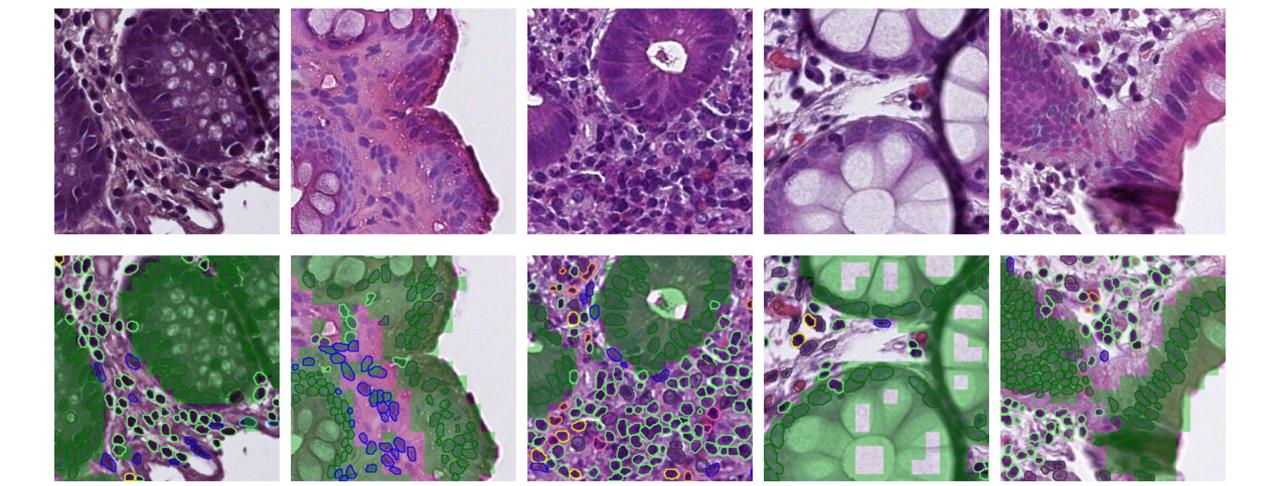}
    \end{tabular}
  \caption{Max (top two rows) and min (bottom two rows) tiles within the IBDColEpi cohort, with overlays of the predictions from HistoPLUS and EpiSeg.}
  \label{fig:min_max_ibdcolepi}
\end{figure}

\end{document}